\documentclass{article}%
\usepackage{amsmath}
\usepackage{amsfonts}
\usepackage{amssymb}
\usepackage{graphicx}
\usepackage{algorithmic}
\usepackage{algorithm}%
\providecommand{\U}[1]{\protect\rule{.1in}{.1in}}

\begin{document}

\title{Attention-based Random Forest and Contamination Model}
\author{Lev V. Utkin and Andrei V. Konstantinov\\Peter the Great St.Petersburg Polytechnic University\\St.Petersburg, Russia\\e-mail: lev.utkin@gmail.com, andrue.konst@gmail.com}
\date{}
\maketitle

\begin{abstract}
A new approach called ABRF (the attention-based random forest) and its
modifications for applying the attention mechanism to the random forest (RF)
for regression and classification are proposed. The main idea behind the
proposed ABRF models is to assign attention weights with trainable parameters
to decision trees in a specific way. The weights depend on the distance
between an instance, which falls into a corresponding leaf of a tree, and
instances, which fall in the same leaf. This idea stems from representation of
the Nadaraya-Watson kernel regression in the form of a RF. Three modifications
of the general approach are proposed. The first one is based on applying the
Huber's contamination model and on computing the attention weights by solving
quadratic or linear optimization problems. The second and the third
modifications use the gradient-based algorithms for computing trainable
parameters. Numerical experiments with various regression and classification
datasets illustrate the proposed method.


\end{abstract}

\section{Introduction}

The attention mechanism can be regarded as a tool by which a neural network
can automatically distinguish the relative importance of features or instances
and weigh them for improving the classification or regression accuracy. It can
be viewed as a learnable mask which emphasizes relevant information in a
feature map. Originally, attention stems from a property of the human
perception to selectively concentrate on an important part of information and
to ignore other information \cite{Niu-Zhong-Yu-21}. Therefore, many
applications of the attention mechanism focus on the natural language
processing (NLP) models, including text classification, translation, etc. Many
attention-based models are also applied to the computer vision area, including
image-based analysis, visual question answering, etc. Detailed surveys and
reviews of attention, its forms, properties and applications can be found in
\cite{Chaudhari-etal-2019,Correia-Colombini-21a,Correia-Colombini-21,Lin-Wang-etal-21,Niu-Zhong-Yu-21}%
.

Many neural attention models are simple, scalable, flexible, and with
promising results in several application domains \cite{Correia-Colombini-21a}.
However, attention is considered as an essential component of neural
architectures \cite{Chaudhari-etal-2019}, the attention weights are learned by
incorporating an additional feed forward neural network within the
architectures. This implies that they have all problems which are encountered
in neural networks, including, overfitting, many tuning parameters,
requirements of a large amount of data, the black-box nature, expensive
computations. One of the powerful models different from neural networks is the
random forest (RF) \cite{Breiman-2001}, which uses a large number of randomly
built individual decision trees in order to combine their predictions. RFs
reduce the possible correlation between decision trees by selecting different
subsamples of the feature space and different subsamples of instances.

Our aim is to avoid using neural networks and to propose the attention-based
random forest (ABRF) models which can be regarded as an efficient alternative
to neural networks in many applications. Moreover, the ABRF models can also be
an efficient alternative to the RF and enhance the RF regression and
classification accuracy.

The main idea behind the proposed ABRF models is to assign weights to decision
trees in a specific way. The weights can be regarded as the attention weights
because they are defined by using queries, keys and values concepts in terms
of the attention mechanism. In contrast to weights of trees defined in
\cite{Utkin-Kovalev-Coolen-2020,Utkin-Kovalev-Meldo-2019}, weights in the ABRF
have trainable parameters and depend on how far an instance, which falls into
a leaf, is from instances, which fall in the same leaf. The resulting
prediction of the ABRF is computed as a weighted sum of predictions obtained
by means of decision trees.

It is pointed out in \cite{Chaudhari-etal-2019,Zhang2021dive} that the
original idea of attention can be understood from the statistics point of view
applying the Nadaraya-Watson kernel regression model
\cite{Nadaraya-1964,Watson-1964}. According to the Nadaraya-Watson regression,
weights or normalized kernels conform with relevance of a training instance to
a target feature vector. Our idea in terms of the original attention mechanism
is to consider every decision tree prediction as the value, the training
instance as the key, the testing instance as the query. In fact, we combine
the Nadaraya-Watson kernel regression model in the form of the RF. The most
interesting question with respect to this representation is how to compute and
to train weights of trees, how to define trainable parameters of the weights.

Depending on a way for the definition of weights and a way for their training,
we propose the following three modifications of the ABRF models:

\begin{enumerate}
\item \textbf{ABRF-1} whose decision tree weights are learned by using the
well-known Huber's $\epsilon$-contamination model \cite{Huber81} where the
trainable parameters of weights are optimally selected from an arbitrarily
adversary distribution. Each weight consists of two parts: the softmax
operation with the tuning coefficient $1-\epsilon$ and the trainable bias of
the softmax weight with coefficient $\epsilon$. This is a very surprising link
between robust contamination model and the attention mechanism. An advantage
of using the $\epsilon$-contamination model for training the attention weights
is that the tree weights used to form prediction are linear on trainable
parameters. This fact allows us to formulate optimization problem in terms of
quadratic or even linear programming, i.e., we get the linear or quadratic
programming problems for computing optimal weights and do not use the
gradient-based algorithm for learning. The quadratic programming problem has a
unique solution, and there are many efficient algorithms for its solving. 

\item \textbf{ABRF-2} whose decision tree weights are defined in the standard
way through the softmax operation where the trainable parameters are
incorporated into the softmax function. This modification is more flexible
because trainable parameters and the entire attention weights may be defined
in various ways, for example, in conventional ways accepted in attention
models \cite{Bahdanau-etal-14,Luong-etal-2015,Niu-Zhong-Yu-21,Vaswani-etal-17}%
. In spite of the flexibility of the modification, it requires using gradient
methods for searching for optimal weights or their trainable parameters with
all virtues and shortcomings of the computational methods.

\item \textbf{ABRF-3} is a combination of ABRF-1 and ABRF-2. The modification
extends the set of trainable parameters of attention weights. We learn two
subsets of parameters. The first one consists of the trainable biases in the
$\epsilon$-contamination model, the second subset includes parameters
incorporated into the softmax operation. It is obvious that gradient descent
methods can be used for all parameters. The third modification can be regarded
as a generalization of ABRF-1 and ABRF-2.
\end{enumerate}

One of the important advantages of the ABRF (ABRF-1) is that the simple
quadratic or linear optimization problems have to be solved to train weights
of trees. Another advantage is that we do not need to rebuild trees after
appearing new training examples. It is enough to train new weights of trees.
The models provide interpretable predictions because weights of trees show
which decision trees have the largest contribution into predictions. There are
also other advantages which will be pointed out below. To the best of our
knowledge, there are no similar attention-based RF models combined with the
Huber's $\epsilon$-contamination model. Due to their flexibility, they can be
a basis for development quite a new class of attention-based models trained
especially on small datasets when the original neural network models may
provide worse results.

Many numerical experiments are provided for studying the proposed
attention-based models. We investigate two types of RFs: original RFs and
Extremely Randomized Trees (ERT). At each node, the ERT algorithm chooses a
split point randomly for each feature and then selects the best split among
these \cite{Geurts-etal-06}.

The paper is organized as follows. Related work can be found in Section 2. A
brief introduction to the attention mechanism is given in Section 3. The
proposed model and its modifications for regression are provided in Section 4.
Some changes of the general model for classification are considered in Section
5. Numerical experiments illustrating regression and classification problems
are provided in Section 6. Concluding and discussion remarks can be found in
Section 7.

\section{Related work}

\textbf{Attention mechanism}. Many attention-based models have been developed
to improve the performance of classification and regression algorithms.
Surveys of various attention-based models are available in
\cite{Chaudhari-etal-2019,Correia-Colombini-21a,Correia-Colombini-21,Lin-Wang-etal-21,Liu-Huang-etal-21,Niu-Zhong-Yu-21}%
.

It should be noted that one of the computational problems of attention
mechanisms is training the softmax function. In order to overcome this
difficulty, several interesting approaches have been proposed. Choromanski et
al. \cite{Choromanski-etal-21} introduced Performers as a Transformer
architecture which can estimate softmax attention with provable accuracy using
only linear space and time complexity. A linear unified nested attention
mechanism that approximates softmax attention with two nested linear attention
functions was proposed by Ma et al. \cite{Ma-Kong-etal-21}. A new class of
random feature methods for linearizing softmax and Gaussian kernels called
hybrid random features (HRFs) was introduced in \cite{Choromanski-etal-21a}.
The same problem is solved in \cite{Peng-Pappas-etal-21} where the authors
propose random feature attention, a linear time and space attention that uses
random feature methods to approximate the softmax function. Schlag et al.
\cite{Schlag-etal-2021} proposed a new kernel function to linearize attention
which balances simplicity and effectiveness. A detailed survey of techniques
of random features to speed up kernel methods was provided by Liu et al.
\cite{Liu-Huang-etal-21}.

We propose three modifications of the attention-based RF such that the second
and the third modifications (ABRF-2 and ABRF-3) have trainable softmax
functions and meet computational problem to solve the corresponding
optimization problems. However, the first modification (ABRF-1) is based on
solving the standard linear or quadratic optimization problems. It overcomes
the computational problems of the attention mechanism.

\textbf{Weighted RFs}. Various approaches were developed to implement the
weighted random forests. Approaches of the first group are based on assigning
weights to decision trees in accordance with some criteria to improve the
classification and regression models
\cite{Kim-Kim-Moon-Ahn-2011,Li-Wang-Ding-Dong-2010,Ronao-Cho-2015,Winham-etal-2013,Xuan-etal-18,Zhang-Wang-21}%
. There are approaches \cite{Daho-2014} which use weights of classes to deal
with imbalanced datasets. However, the assigned weights in the aforementioned
works are not trainable parameters. Attempts to train weights of trees were
carried out in
\cite{Utkin-Konstantinov-etal-20,Utkin-etal-2019,Utkin-Kovalev-Coolen-2020,Utkin-Kovalev-Meldo-2019}%
, where weights are assigned by solving optimization problems, i.e., they
incorporated into a certain loss function of the whole RF such that the loss
function is minimized over values of weights.

In contrast to the above approaches, weights assigned to trees in the proposed
attention-based models depend not only on trees, but on each instance, i.e.,
vectors of the feature weights, the preliminary tree weights, the instance
weights are introduced and learned to implement the attention mechanism. This
is a very important difference of the proposed attention-based models from the
available weighted RFs.

\section{Preliminary: The attention mechanism}

The attention mechanism can be regarded as a tool by which a neural network
can automatically distinguish the relative importance of features and weigh
the features for enhancing the classification accuracy. It can be viewed as a
learnable mask which emphasizes relevant information in a feature map. It is
pointed out in \cite{Chaudhari-etal-2019,Zhang2021dive} that the original idea
of attention can be understood from the statistics point of view applying the
Nadaraya-Watson kernel regression model \cite{Nadaraya-1964,Watson-1964}.

Given $n$ instances $S=\{(\mathbf{x}_{1},y_{1}),(\mathbf{x}_{2},y_{2}%
),...,(\mathbf{x}_{n},y_{n})\}$, in which $\mathbf{x}_{i}=(x_{i1}%
,...,x_{im})\in\mathbb{R}^{m}$ represents a feature vector involving $m$
features and $y_{i}\in\mathbb{R}$ represents the regression outputs, the task
of regression is to construct a regressor $f:\mathbb{R}^{m}\rightarrow
\mathbb{R}$ which can predict the output value $\tilde{y}$ of a new
observation $\mathbf{x}$, using available data $S$. The similar task can be
formulated for the classification problem.

The original idea behind the attention mechanism is to replace the simple
average of outputs $\tilde{y}=n^{-1}\sum_{i=1}^{n}y_{i}$ for estimating the
regression output $y$, corresponding to a new input feature vector
$\mathbf{x}$ with the weighted average, in the form of the Nadaraya-Watson
regression model \cite{Nadaraya-1964,Watson-1964}:%
\begin{equation}
\tilde{y}=\sum_{i=1}^{n}\alpha(\mathbf{x},\mathbf{x}_{i})y_{i},
\end{equation}
where weight $\alpha(\mathbf{x},\mathbf{x}_{i})$ conforms with relevance of
the $i$-th instance to the vector $\mathbf{x}$.

According to the Nadaraya-Watson regression model, to estimate the output $y$
of an input variable $\mathbf{x}$, training outputs $y_{i}$ given from a
dataset weigh in agreement with the corresponding input $\mathbf{x}_{i}$
locations relative to the input variable $\mathbf{x}$. The closer an input
$\mathbf{x}_{i}$ to the given variable $\mathbf{x}$, the greater the weight
assigned to the output corresponding to $\mathbf{x}_{i}$.

One of the original forms of weights is defined by a kernel $K$ (the
Nadaraya-Watson kernel regression \cite{Nadaraya-1964,Watson-1964}), which can
be regarded as a scoring function estimating how vector $\mathbf{x}_{i}$ is
close to vector $\mathbf{x}$. The weight is written as follows:
\begin{equation}
\alpha(\mathbf{x},\mathbf{x}_{i})=\frac{K(\mathbf{x},\mathbf{x}_{i})}%
{\sum_{j=1}^{n}K(\mathbf{x},\mathbf{x}_{j})}.
\end{equation}

In terms of the attention mechanism \cite{Bahdanau-etal-14}, vector
$\mathbf{x}$, vectors $\mathbf{x}_{i}$ and outputs $y_{i}$ are called as the
\textit{query}, \textit{keys} and \textit{values}, respectively. Weight
$\alpha(\mathbf{x},\mathbf{x}_{i})$ is called as the attention weight.
Generally, weights $\alpha(\mathbf{x},\mathbf{x}_{i})$ can be extended by
incorporating trainable parameters. For example, if we denote $\mathbf{q=W}%
_{q}\mathbf{x}$ and $\mathbf{k}_{i}\mathbf{=W}_{k}\mathbf{x}_{i}$ referred to
as the query and key embeddings, respectively, then the attention weight can
be represented as:%

\begin{equation}
\alpha(\mathbf{x},\mathbf{x}_{i})=\text{\textrm{softmax}}\left(
\mathbf{q}^{\mathrm{T}}\mathbf{k}_{i}\right)  =\frac{\exp\left(
\mathbf{q}^{\mathrm{T}}\mathbf{k}_{i}\right)  }{\sum_{j=1}^{n}\exp\left(
\mathbf{q}^{\mathrm{T}}\mathbf{k}_{j}\right)  }, \label{Expl_At_12}%
\end{equation}
where $\mathbf{W}_{q}$ and $\mathbf{W}_{k}$ are matrices of parameters which
are learned, for example, by incorporating an additional feed forward neural
network within the system architecture.

There exist several definitions of attention weights and the corresponding
attention mechanisms, for example, the additive attention
\cite{Bahdanau-etal-14}, multiplicative or dot-product attention
\cite{Luong-etal-2015,Vaswani-etal-17}. We propose a new attention mechanism
which is based on the weighted RFs training and the Huber's $\epsilon
$-contamination model.

\section{Attention-based RF: Regression models}

Let us formally state the standard regression problem. Given $n$ training data
(examples, instances, patterns) $S=\{(\mathbf{x}_{1},y_{1}),...,(\mathbf{x}%
_{n},y_{n})\}$, in which $\mathbf{x}_{i}$ may belong to an arbitrary set
$\mathcal{X}\subset\mathbb{R}^{m}$ and represents a feature vector involving
$m$ features, and $y_{i}\in\mathbb{R}$ represents the observed output such
that $y_{i}=f(\mathbf{x}_{i})+\xi$. Here $\xi$ is the random noise with
expectation $0$ and a finite variance. Machine learning aims to construct a
regression model $f(\mathbf{x})$ that minimizes the expected risk which, for
example, can be represented by the least squares error criterion
\begin{equation}
\frac{1}{n}\sum_{i=1}^{n}\left(  y_{i}-f(\mathbf{x}_{i})\right)
^{2}.\label{Impr_DF_100}%
\end{equation}

RFs can be regarded as a powerful nonparametric statistical method for solving
both regression and classification problems. Suppose that a RF consists of $T$
decision trees. Let us write a set of leaves belonging to the $k$-th tree as
$\mathcal{T}^{(k)}=\{t_{1}^{(k)},...,t_{s_{k}}^{(k)}\}$, where $t_{i}^{(k)}$
is the $i$-th leaf in the $k$-th tree; $s_{k}$ is the number of leaves in the
$k$-th tree. Denote indices of instances, which fall into leaf $t_{i}^{(k)}$
as $\mathcal{J}_{i}^{(k)}$ such that $\mathcal{J}_{i}^{(k)}\cap\mathcal{J}%
_{j}^{(k)}=\varnothing$. The last condition is very important and means that
the same instance cannot fall into different leaves of the same tree. Suppose
that an instance $\mathbf{x}$ falls into the $i$-th leaf, i.e., into leaf
$t_{i}^{(k)}$. Let us also introduce the mean vector $\mathbf{A}%
_{k}(\mathbf{x)}$ and the mean target value $B_{k}$ defined as the mean of
training instance vectors, which fall into the $i$-th leaf of the $k$-th tree,
and the corresponding observed outputs, respectively, i.e.,
\begin{equation}
\mathbf{A}_{k}(\mathbf{x)}=\frac{1}{\#\mathcal{J}_{i}^{(k)}}\sum
_{j\in\mathcal{J}_{i}^{(k)}}\mathbf{x}_{j}, \label{RF_Att_20}%
\end{equation}%
\begin{equation}
B_{k}(\mathbf{x)}=\frac{1}{\#\mathcal{J}_{i}^{(k)}}\sum_{i\in\mathcal{J}%
_{j}^{(k)}}y_{j}. \label{RF_Att_21}%
\end{equation}

We do not use index $i$ of the leaf in notations of $\mathbf{A}_{k}%
(\mathbf{x)}$ and $B_{k}(\mathbf{x)}$ because instance $\mathbf{x}$ can fall
only into one leaf. It should be noted that $B_{k}(\mathbf{x)}$ in regression
trees is nothing else but the $k$-th tree prediction $\tilde{y}_{k}$
corresponding to instance $\mathbf{x}$, i.e., $\tilde{y}=B_{k}(\mathbf{x)}$.
In other words, the predicted value $\tilde{y}$ assigned to leaf $t_{i}%
^{(k)}\in\mathcal{T}^{(k)}$ is the average of outputs of all instances with
indices from $\mathcal{J}_{i}^{(k)}$. This implies that the predicted value
strongly depends on instances from $\mathcal{J}_{i}^{(k)}$.

Suppose that we have trained trees in the RF consisting of $T$ trees. If an
instance $\mathbf{x}$ (training or testing) falls into leaf $t_{i}^{(k)}%
\in\mathcal{T}^{(k)}$ of the $k$-th tree, then the distance $d\left(
\mathbf{x},\mathbf{A}_{k}(\mathbf{x)}\right)  $ shows how far instance
$\mathbf{x}$ is from the mean vector of all instances which fall into leaf
$t_{i}^{(k)}$. We use the $L_{2}$-norm for the distance definition, i.e.,
$d\left(  \mathbf{x},\mathbf{A}_{k}(\mathbf{x)}\right)  =\left\Vert
\mathbf{x}-\mathbf{A}_{k}(\mathbf{x)}\right\Vert ^{2}$. Note that each tree
has only a single leaf where an instance falls into.

According to the standard regression RF, the final RF prediction $\tilde{y}$
for a testing instance $\mathbf{x}$ is defined as
\begin{equation}
\tilde{y}=\frac{1}{T}\sum_{k=1}^{T}\tilde{y}_{k}=\frac{1}{T}\sum_{k=1}%
^{T}B_{k}(\mathbf{x)}.
\end{equation}

The above prediction assumes that all trees have the same contribution into
the final prediction $\tilde{y}$, their weights are identical and equal to
$1/T$. However, one can see that distances between $\mathbf{x}$ and
$\mathbf{A}_{k}(\mathbf{x)}$ are different and may differently impact on
$\tilde{y}$. Therefore, we can introduce weights of trees such that a larger
weight is assigned to a tree with the smaller distance because these trees
provide better accuracy and vice versa. The next question is how to define the
weights of trees.

Let us return to the definition of the Nadaraya-Watson regression model and
rewrite it in terms of the RF as follows:
\begin{equation}
\tilde{y}=\sum_{k=1}^{T}\alpha\left(  \mathbf{x},\mathbf{A}_{k}(\mathbf{x)}%
,\mathbf{w}\right)  \cdot\tilde{y}_{k}=\sum_{k=1}^{T}\alpha\left(
\mathbf{x},\mathbf{A}_{k}(\mathbf{x)},\mathbf{w}\right)  \cdot B_{k}%
(\mathbf{x)}. \label{RF_Att_47}%
\end{equation}

Here $\alpha\left(  \mathbf{x},\mathbf{A}_{k}(\mathbf{x)},\mathbf{w}\right)  $
is the attention weight which conforms with relevance of \textquotedblleft
mean instance\textquotedblright\ $\mathbf{A}_{k}(\mathbf{x)}$ to vector
$\mathbf{x}$ and satisfies condition
\begin{equation}
\sum_{k=1}^{T}\alpha\left(  \mathbf{x},\mathbf{A}_{k}(\mathbf{x)}%
,\mathbf{w}\right)  =1,
\end{equation}
$\mathbf{w}$ is a vector of training attention parameters which will be
defined below in accordance with the model modification. In terms of the
attention mechanism for the $k$-th tree, $\tilde{y}_{k}$ or $B_{k}%
(\mathbf{x)}$ is the \textit{value}, $\mathbf{A}_{k}(\mathbf{x)}$ is the
\textit{key}, and $\mathbf{x}$ is the \textit{query}.

In sum, we get the trainable attention-based RF with parameters $\mathbf{w}$
which are defined by minimizing the expected loss function over a set
$\mathcal{W}$ of parameters as follows:
\begin{equation}
\mathbf{w}_{opt}=\arg\min_{\mathbf{w\in}\mathcal{W}}~\sum_{s=1}^{n}L\left(
\tilde{y}_{s},y_{s},\mathbf{w}\right)  . \label{RF_Att_49}%
\end{equation}

If we use the $L_{2}$-norm for the distance $d\left(  \mathbf{x}%
_{s},\mathbf{A}_{k}(\mathbf{x}_{s}\mathbf{)}\right)  $ and for the difference
between $\tilde{y}_{s}$ and $y_{s}$, then the loss function can be rewritten
as
\begin{align}
\sum_{s=1}^{n}L\left(  \tilde{y}_{s},y_{s},\mathbf{w}\right)   &  =\sum
_{s=1}^{n}\left(  y_{s}-\sum_{k=1}^{T}\alpha\left(  \mathbf{x}_{s}%
,\mathbf{A}_{k}(\mathbf{x}_{s}\mathbf{)},\mathbf{w}\right)  \cdot\tilde{y}%
_{s}\right)  ^{2}\nonumber\\
&  =\sum_{s=1}^{n}\left(  y_{s}-\sum_{k=1}^{T}\alpha\left(  \left\Vert
\mathbf{x}_{s}-\mathbf{A}_{k}(\mathbf{x}_{s}\mathbf{)}\right\Vert
^{2},\mathbf{w}\right)  B_{k}(\mathbf{x}_{s}\mathbf{)}\right)  ^{2}.
\label{RF_Att_50}%
\end{align}

Our preliminary numerical experiments have shown that usage of the feature
weights may significantly improve the model performance. Therefore, we also
introduce the vector $\mathbf{z}=(z_{1},...,z_{m})\in\mathbb{R}_{+}^{m}$ such
that $\sum_{i=1}^{m}z_{i}=1$, which is regarded as a vector of feature
weights. Hence, (\ref{RF_Att_50}) can be rewritten as
\begin{equation}
\sum_{s=1}^{n}L\left(  \tilde{y}_{s},y_{s},\mathbf{w},\mathbf{z}\right)
=\sum_{s=1}^{n}\left(  y_{s}-\sum_{k=1}^{T}\alpha\left(  \left\Vert \left(
\mathbf{x}_{s}-\mathbf{A}_{k}(\mathbf{x}_{s}\mathbf{)}\right)  \circ
\mathbf{z}\right\Vert ^{2},\mathbf{w}\right)  B_{k}(\mathbf{x}_{s}%
\mathbf{)}\right)  ^{2}. \label{RF_Att_51}%
\end{equation}

Here \textquotedblleft$\circ$\textquotedblright\ means Hadamard product of vectors.

The next question is how to define the function $\alpha$ and how to compute
trainable parameters $\mathbf{w}$ and $\mathbf{z}$. Three modifications of
ABRF correspond to different functions $\alpha$.

\subsection{ABRF-1}

\subsubsection{A case of quadratic programming}

A common way for defining the function $\alpha$ in the attention models is to
use the softmax operation which is given in (\ref{Expl_At_12}). The softmax
function has several desirable properties, for example, the sum of all its
values is equal to $1$. However, if we incorporate trainable parameters
$\mathbf{w}$ into the function as it is shown in (\ref{Expl_At_12}), then
computing optimal values of $\mathbf{w}$ leads to using the gradient-based
methods. In order to simplify computations and to get a unique solution for
$\mathbf{w}$, we would like to avoid using the gradient-based algorithms. To
achieve this goal, we propose to apply the well-known Huber's $\epsilon
$-contamination model \cite{Huber81} where the trainable parameters of weights
are optimally selected from an arbitrarily adversary distribution. The
$\epsilon$-contamination model can be represented as
\begin{equation}
(1-\epsilon)\cdot P+\epsilon\cdot Q, \label{RF_Att_40}%
\end{equation}
where the probability distribution $P$ is contaminated by some arbitrary
distribution $Q$; the rate $\epsilon\in\lbrack0,1]$ is a model parameter which
reflects how \textquotedblleft close\textquotedblright\ we feel that $Q$ must
be to $P$ \cite{Berger85}.

Usage of the $\epsilon$-contamination model stems from several reasons. First,
the softmax function can be interpreted as the probability distribution $P$ in
(\ref{RF_Att_40}) because its sum is $1$. It is a point in the probabilistic
unit simplex having $T$ vertices. Second, weights $\alpha\left(
\mathbf{x}_{s},\mathbf{A}_{k}(\mathbf{x}_{s}\mathbf{)},\mathbf{w}\right)  $
also can be interpreted as a probability distribution or another point in the
same unit simplex. This point is biased by means of the third probability
distribution $Q$ in (\ref{RF_Att_40}) which is trained in order to achieve the
best prediction results. The contamination parameter $\epsilon$ can be
regarded as a tuning or training parameter of the model. Hereinafter, we
consider it only as a tuning parameter. Hence, we can define the attention
weights $\alpha$ as follows:
\begin{equation}
\alpha\left(  \mathbf{x}_{s},\mathbf{A}_{k}(\mathbf{x}_{s}\mathbf{)}%
,\mathbf{w}\right)  =(1-\epsilon)\cdot\text{\textrm{softmax}}\left(
d(\mathbf{x}_{s}\mathbf{A}_{k}(\mathbf{x}_{s}\mathbf{))}\right)
+\epsilon\cdot w_{k}.\label{RF_Att_41}%
\end{equation}

By using the expected loss function (\ref{RF_Att_50}) and the definition of
the attention weights (\ref{RF_Att_41}), we can write the following quadratic
optimization problem for computing optimal distributions $\mathbf{w}$:
\begin{equation}
\min_{\mathbf{w\in}\mathcal{W}}\sum_{s=1}^{n}\left(  y_{s}-\sum_{k=1}%
^{T}\left(  (1-\epsilon)D_{k}(\mathbf{x}_{s},\tau)+\epsilon w_{k}\right)
\cdot B_{k}(\mathbf{x}_{s}\mathbf{)}\right)  ^{2}, \label{RF_Att_42}%
\end{equation}
subject to $w_{k}\geq0$, $k=1,...,T$, and $\sum_{k=1}^{T}w_{k}=1$,

Here
\begin{equation}
D_{k}(\mathbf{x}_{s},\tau)=\text{\textrm{softmax}}\left(  \frac{\left\Vert
\mathbf{x}_{s}-\mathbf{A}_{k}(\mathbf{x}_{s}\mathbf{)}\right\Vert ^{2}}{2\tau
}\right)  , \label{RF_Att_43}%
\end{equation}
where $\tau$ is a tuning parameter (temperature) of the softmax function; set
$\mathcal{W}$ is the unit simplex.

Note that we do not use the vector of feature weights $\mathbf{z}$ in ABRF-1
to have a quadratic optimization problem and to avoid the gradient-based
algorithms. We also should point out that $D_{k}(\mathbf{x}_{s},\tau)$ in
ABRF-1 does not have trainable parameters. On the one hand, this peculiarity
simplifies the computation problem. On the other hand, it reduces the strength
of the attention mechanism.

\begin{figure}
[ptb]
\begin{center}
\includegraphics[
height=1.8299in,
width=3.1073in
]%
{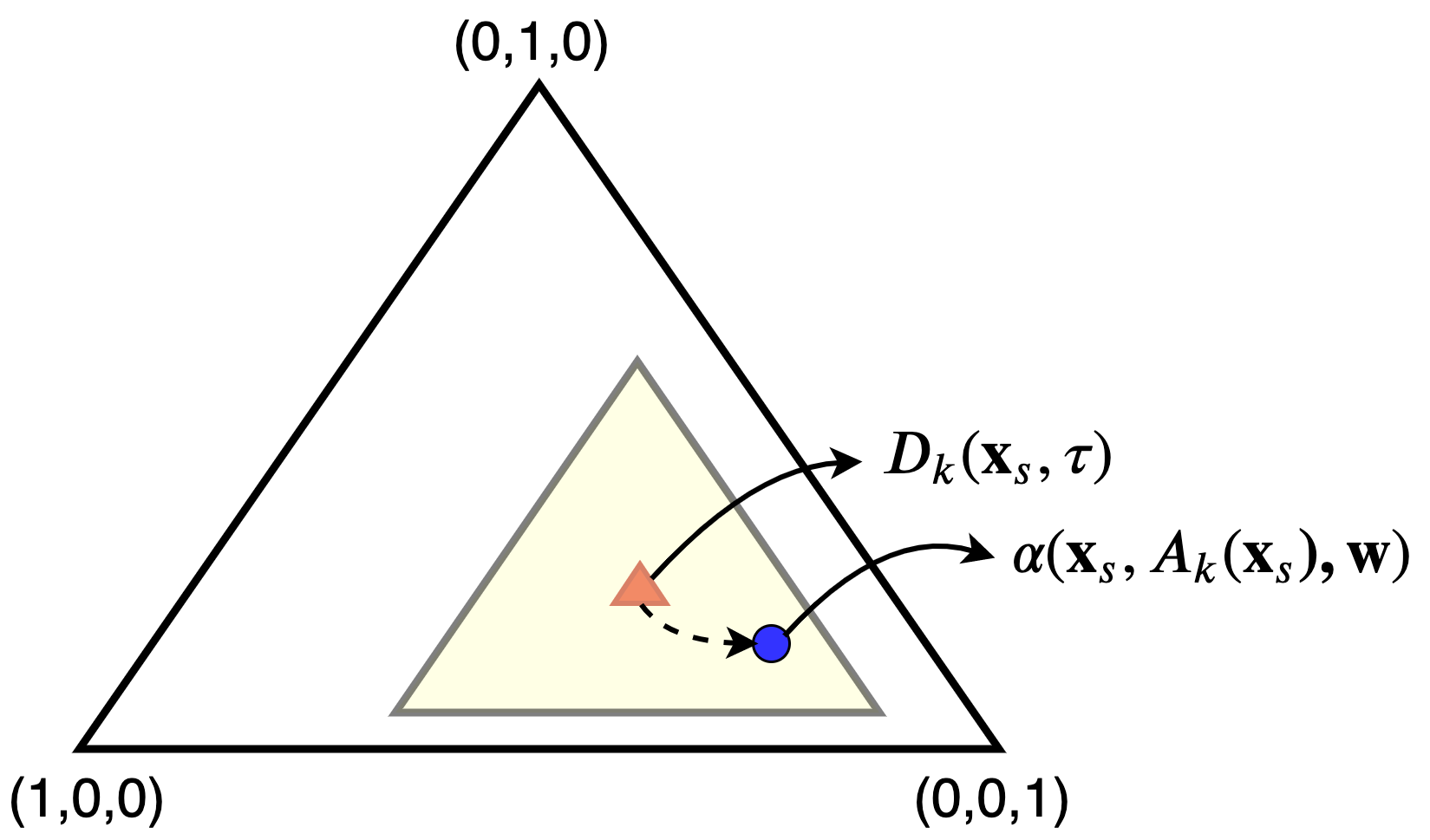}%
\caption{The unit simplex with points $D_{k}(\mathbf{x}_{s},\tau)$ (the small
triangle) and $\alpha\left(  \mathbf{x}_{s},A_{k}(\mathbf{x}_{s}%
\mathbf{)},\mathbf{w}\right)  $ (the small circle)}%
\label{f:RF_Att_1}%
\end{center}
\end{figure}

Let us consider the attention weights $\alpha\left(  \mathbf{x},\mathbf{A}%
_{k}(\mathbf{x)},\mathbf{w}\right)  $ produced by the sum $(1-\epsilon
)D_{k}(\mathbf{x}_{s},\tau)+\epsilon w_{k}$ in detail. If $\mathbf{D}%
_{s}=(D_{1}(\mathbf{x}_{s},\tau),...,D_{T}(\mathbf{x}_{s},\tau))$ is a point
in the unit simplex, then $\alpha\left(  \mathbf{x}_{s},\mathbf{A}%
_{k}(\mathbf{x}_{s}\mathbf{)},\mathbf{w}\right)  $ belongs to a subsimplex
whose size depends on the value of $\epsilon$. The unit simplex for the case
$T=3$ (three decision trees) with vertices $(1,0,0)$, $(0,1,0)$, $(0,0,1)$ is
depicted in Fig. \ref{f:RF_Att_1} where the small triangle and the small
circle are points $\mathbf{D}_{s}$ and $\alpha\left(  \mathbf{x}%
_{s},\mathbf{A}_{k}(\mathbf{x}_{s}\mathbf{)},\mathbf{w}\right)  $,
respectively. It can be seen from Fig. \ref{f:RF_Att_1} that the attention
weights are located in the subsimplex of a smaller size. The corresponding
point is biased by the vector $\mathbf{w}$ whose optimal value is computed by
solving the quadratic optimization problem (\ref{RF_Att_42}). The attention
weight point cannot be outside the small subsimplex whose center is
$\mathbf{D}_{s}$, and its size is defined by parameter $\epsilon$. On the one
hand, we should increase $\epsilon$ in order to increase the subsimplex and to
extend the set of solutions. On the other hand, the extension of the solution
set may lead to the sparse solution and to reducing the role of instances
because weights of trees $\alpha$ in this case are obtained mainly from
$\mathbf{w}$, and they do not take into account relationship between
$\mathbf{x}$ and $\mathbf{A}_{k}(\mathbf{x)}$. The optimal choice of
$\epsilon$ is carried out by testing ABRF-1 for different values of $\epsilon$.

The first advantage of ABRF-1 modification is that it is simple from the
computational point of view because its training is based on solving the
standard quadratic optimization problem. The second advantage is that if
$\mathbf{D}_{s}$ is a probability distribution, then, due to the $\epsilon
$-contamination model the attention weights also compose a probability
distribution. This implies that we do not need to worry about non-negative
values of weights and the unit sum of weights. The third advantage of ABRF-1
is that it is simply interpreted, i.e., the obtained weights indicate which
decision trees significantly contribute into the RF prediction.%

\subsubsection{A case of linear programming}

So far, we have studied the case of $L_{2}$-norm for defining the loss
function $L\left(  \tilde{y}_{s},y_{s},\mathbf{w}\right)  $ in
(\ref{RF_Att_50}). This norm leads to the quadratic optimization problem in
the ABRF-1 model. However, we can also consider the $L_{1}$-norm. It turns out
that this norm leads to the linear optimization problem. Denote for
simplicity
\begin{equation}
Q_{s}=y_{s}-(1-\epsilon)\sum_{k=1}^{T}D_{k}(\mathbf{x}_{s},\tau)B_{k}%
(\mathbf{x}_{s}\mathbf{).}%
\end{equation}

Then we can rewrite optimization problem (\ref{RF_Att_42}) by using the
$L_{1}$-norm as follows:%

\begin{equation}
\min_{\mathbf{w\in}\mathcal{W}}\sum_{s=1}^{n}\left\vert Q_{s}-\epsilon
\sum_{k=1}^{T}B_{k}(\mathbf{x}_{s}\mathbf{)}w_{k}\right\vert .
\end{equation}

Introduce the following optimization variables:%

\begin{equation}
G_{s}=\left\vert Q_{s}-\epsilon\sum_{k=1}^{T}B_{k}(\mathbf{x}_{s}%
\mathbf{)}w_{k}\right\vert ,\ s=1,...,n.
\end{equation}

Then we get the linear optimization problem:%

\begin{equation}
\min_{\mathbf{w},G_{1},...,G_{n}}\sum_{s=1}^{n}G_{s}, \label{RF_Att_34}%
\end{equation}
subject to $w_{k}\geq0$, $k=1,...,T$, and $\sum_{k=1}^{T}w_{k}=1$,
\begin{equation}
G_{s}\geq Q_{s}-\epsilon\sum_{k=1}^{T}B_{k}(\mathbf{x}_{s}\mathbf{)}%
w_{k},\ s=1,...,n, \label{RF_Att_35}%
\end{equation}%
\begin{equation}
G_{s}\geq-Q_{s}+\epsilon\sum_{k=1}^{T}B_{k}(\mathbf{x}_{s}\mathbf{)}%
w_{k},\ s=1,...,n. \label{RF_Att_36}%
\end{equation}

The above problem has $n+T$ variables and $2n+T+1$ linear constraints. Its
solution does not meet any difficulties.

\subsection{ABRF-2}

Another modification (ABRF-2) is based on using conventional definitions of
the attention scoring function when trainable parameters are incorporated into
the score function
\cite{Bahdanau-etal-14,Luong-etal-2015,Niu-Zhong-Yu-21,Vaswani-etal-17}. One
of the simplest way for defining the attention weights is to apply
(\ref{Expl_At_12}). Then we get the following weight for the $k$-th tree:
\begin{equation}
\alpha\left(  \mathbf{x},\mathbf{A}_{k}(\mathbf{x)},\mathbf{v},\mathbf{z}%
\right)  =\text{\textrm{softmax}}\left(  \frac{\left\Vert \left(
\mathbf{x}-\mathbf{A}_{k}(\mathbf{x)}\right)  \circ\mathbf{z}\right\Vert ^{2}%
}{2}v_{k}\right)  ,\ k=1,...,T, \label{RF_Att_48}%
\end{equation}
where $\mathbf{v}=(v_{1},...,v_{T})\in\mathbb{R}_{+}^{T}$ is the vector of
training parameters such that $\sum_{k=1}^{T}v_{k}=1$; $\mathbf{z}%
=(z_{1},...,z_{m})\in\mathbb{R}_{+}^{m}$ is the training vector of the feature weights.

In fact, we replace the vector of parameters $\mathbf{w}$ by the vector
$\mathbf{v}$ and the vector $\mathbf{z}$. It should be noted that the
temperature parameter $\tau$ is no longer used because it is incorporated into
training parameters $\mathbf{v}$. In this case, optimization problem
(\ref{RF_Att_51}) with (\ref{RF_Att_48}) cannot be represented as the
quadratic optimization problem. However, optimal values of parameters
$\mathbf{v}$ and $\mathbf{z}$ can be found by means of well-known optimization
methods, for example, by means of the gradient descent algorithm. It should be
also noted that only one of the vectors $\mathbf{z}$ and $\mathbf{v}$ can be
used as trainable parameters of the attention mechanism. However, our
numerical experiments show that both vectors $\mathbf{z}$ and $\mathbf{v}$
provide outperforming results.

\subsection{ABRF-3}

The third modification (ABRF-3) combines ABRF-1 and ABRF-2 models, i.e., it
uses three vectors of training parameters $\mathbf{v}$, $\mathbf{z}$ and
$\mathbf{w}$. Vectors $\mathbf{v}$ and $\mathbf{z}$ are incorporated into the
softmax operation (see ABRF-2), vector $\mathbf{w}$ forms the $\epsilon
$-contamination model. As a result, we get the optimization problem whose
objective function coincides with (\ref{RF_Att_42}), but it is minimized over
unit simplices produced by vectors $\mathbf{v}$ and $\mathbf{w}$. Weights of
trees are defined now as follows:%
\begin{equation}
\alpha\left(  \mathbf{x},\mathbf{A}_{k}(\mathbf{x)},\mathbf{w},\mathbf{v}%
,\mathbf{z}\right)  =(1-\epsilon)\cdot\text{\textrm{softmax}}\left(
\frac{\left\Vert \left(  \mathbf{x}-\mathbf{A}_{k}(\mathbf{x)}\right)
\circ\mathbf{z}\right\Vert ^{2}}{2}v_{k}\right)  +\epsilon w_{k}.
\end{equation}

Substituting the above expression into (\ref{RF_Att_50}), we get the objective
function of the optimization problem with obvious constraints:
\begin{equation}
w_{k},v_{k}\geq0,k=1,...,T,\sum_{k=1}^{T}w_{k}=1,\sum_{k=1}^{T}v_{k}=1,
\end{equation}%
\begin{equation}
z_{i}\geq0,i=1,...,m,\sum_{i=1}^{m}z_{i}=1
\end{equation}

The obtained optimization problem can be solved by means of the gradient
descent algorithm.

\section{Attention-based RF: Classification models}

After studying the attention-based RF for regression, the similar machine
learning problem can be formally stated for classification. In this case, the
output $y_{i}\in\mathcal{Y}=\{1,...,C\}$ represents the class of the
associated instance. The task of classification is to construct an accurate
classifier $c:$ $\mathcal{X}\rightarrow\mathcal{Y}$ which can predict the
unknown class label $y$ of a new observation $\mathbf{x}$, using available
training data, such that it minimizes the classification expected risk.

An important property of decision trees is a probability distribution of
classes determined at each leaf node. This probability distribution can be
used for computing probabilities of classes for the whole RF and for making
decision about a class label of a testing instance. Suppose that $n_{i}^{(k)}$
training instances fall into leaf node $t_{i}^{(k)}$ such that $n_{i}%
^{(k)}(1),...,n_{i}^{(k)}(C)$ vectors belong to classes $1,...,C$,
respectively. Here there holds $n_{i}^{(k)}(1)+,...+n_{i}^{(k)}(C)=n_{i}%
^{(k)}$. Then the class probability distribution for example $\mathbf{x}$,
which falls into leaf node $t_{i}^{(k)}$, is $\mathbf{p}_{k}(\mathbf{x}%
)=(p_{k}(\mathbf{x},1),...,p_{k}(\mathbf{x},C))$, and it can be computed as
$p_{k}(\mathbf{x},c)=n_{i}^{(k)}(c)/n_{i}^{(k)}$, $c=1,...,C$. Index $i$ of
the leaf node in notations of $\mathbf{p}_{k}(\mathbf{x})$ is not used again
because instance $\mathbf{x}$ can fall only into one leaf.

Attention-based classification RF differs from the regression problem
statement (see (\ref{RF_Att_47})-(\ref{RF_Att_50})) only by the class
probability distributions $\mathbf{p}_{k}(\mathbf{x})$ instead of the
point-valued target $\tilde{y}_{s}$. If we again denote the tree attention
weight as $\alpha\left(  \mathbf{x},\mathbf{A}_{k}(\mathbf{x)},\mathbf{w}%
\right)  $, then the class probability distribution $\mathbf{p}(\mathbf{x})$
of the whole RF can be written (see (\ref{RF_Att_47})-(\ref{RF_Att_50})) as
follows:
\begin{equation}
\mathbf{p}(\mathbf{x})=\sum_{k=1}^{T}\alpha\left(  \mathbf{x},\mathbf{A}%
_{k}(\mathbf{x)},\mathbf{w}\right)  \mathbf{p}_{k}(\mathbf{x}).
\end{equation}
Hence, the optimization problem for computing trainable vectors $\mathbf{w}$
defining the tree attention weight $\alpha\left(  \mathbf{x},\mathbf{A}%
_{k}(\mathbf{x)},\mathbf{w}\right)  $ is of the form:%
\begin{equation}
\mathbf{w}_{opt}=\arg\min_{\mathbf{w\in}\mathcal{W}}~\sum_{s=1}^{n}L\left(
\mathbf{p}(\mathbf{x}_{s}),\mathbf{h}_{s},\mathbf{w}\right)  ,
\end{equation}
where the loss function is defined as the expected distance between the
one-hot vector $\mathbf{h}_{s}$ and the RF class probability distribution,
i.e., there holds
\begin{equation}
\sum_{s=1}^{n}L\left(  \mathbf{p}(\mathbf{x}_{s}),\mathbf{h}_{s}%
,\mathbf{w}\right)  =\sum_{s=1}^{n}\left\Vert \mathbf{h}_{s}-\sum_{k=1}%
^{T}\alpha\left(  \mathbf{x}_{s},\mathbf{A}_{k}(\mathbf{x}_{s}\mathbf{)}%
,\mathbf{w}\right)  \mathbf{p}(\mathbf{x}_{s})\right\Vert ^{2}.
\label{RF_Att_59}%
\end{equation}

Here the one-hot vector $\mathbf{h}_{s}=(h_{s}(1),...,h_{s}(C))$ has the unit
element with the index corresponding to the class of the $s$-th instance.
Vector $\mathbf{h}_{s}$ is determined from the training set whereas vector
$\mathbf{p}(\mathbf{x}_{s})$ is computed after training decision trees
counting instances from classes, which fall into the corresponding leaf node.

Objective function (\ref{RF_Att_42}) can be rewritten for the classification
case as follows:%

\begin{align}
&  \min_{\mathbf{w\in}\mathcal{W}}\sum_{s=1}^{n}\left\Vert \mathbf{h}_{s}%
-\sum_{k=1}^{T}\left(  (1-\epsilon)D_{k}(\mathbf{x}_{s},\tau)+\epsilon
w_{k}\right)  \mathbf{p}(\mathbf{x}_{s})\right\Vert ^{2}\nonumber\\
&  =\min_{\mathbf{w\in}\mathcal{W}}\sum_{s=1}^{n}\sum_{c=1}^{C}\left(
h_{s}(c)-\sum_{k=1}^{T}\left(  (1-\epsilon)D_{k}(\mathbf{x}_{s},\tau)+\epsilon
w_{k}\right)  p_{k}(\mathbf{x}_{s},c)\right)  ^{2}, \label{RF_Att_60}%
\end{align}
where $D_{k}(\mathbf{x}_{s},\tau)$ is computed in accordance with
(\ref{RF_Att_43}).

One can see from the above that modifications ABRF-1, ABRF-2, ABRF-3 for
classification can be simply reduced to optimization problems which are
similar to the same problems for regression.

\section{Numerical experiments}

\subsection{Regression}

In order to study the proposed approach for solving regression problems, we
apply datasets which are taken from open sources, in particular:  Diabetes can
be found in the corresponding R Packages; Friedman 1, 2 3 are described at
site: https://www.stat.berkeley.edu/\symbol{126}breiman/bagging.pdf;
Regression and Sparse datasets are available in package \textquotedblleft
Scikit-Learn\textquotedblright. The proposed algorithm is evaluated and
investigated also by the following publicly available datasets from the UCI
Machine Learning Repository \cite{Dua:2019}: Wine Red, Boston Housing,
Concrete, Yacht Hydrodynamics, Airfoil. A brief introduction about these data
sets are given in Table \ref{t:regres_datasets} where $m$ and $n$ are numbers
of features and instances, respectively. A more detailed information can be
found from the aforementioned data resources.%

\begin{table}[tbp] \centering
\caption{A brief introduction about the regression data sets}%
\begin{tabular}
[c]{cccc}\hline
Data set & Abbreviation & $m$ & $n$\\\hline
Diabetes & Diabetes & $10$ & $442$\\\hline
Friedman 1 & Friedman 1 & $10$ & $100$\\\hline
Friedman 2 & Friedman 2 & $4$ & $100$\\\hline
Friedman 3 & Friedman 3 & $4$ & $100$\\\hline
Scikit-Learn Regression & Regression & $100$ & $100$\\\hline
Scikit-Learn Sparse Uncorrelated & Sparse & $10$ & $100$\\\hline
UCI Wine red & Wine & $11$ & $1599$\\\hline
UCI Boston Housing & Boston & $13$ & $506$\\\hline
UCI Concrete & Concrete & $8$ & $1030$\\\hline
UCI Yacht Hydrodynamics & Yacht & $6$ & $308$\\\hline
UCI Airfoil & Airfoil & $5$ & $1503$\\\hline
\end{tabular}
\label{t:regres_datasets}%
\end{table}%

We use the coefficient of determination denoted $R^{2}$ and the mean absolute
error (MAE) for the regression evaluation. The greater the value of the
coefficient of determination and the smaller the MAE, the better results we
get. In all tables, we compare $R^{2}$ and the MAE for three cases:

\begin{enumerate}
\item \textbf{RF}: the RF or the ERT without the softmax and without attention model;

\item \textbf{Softmax} model: the RF or the ERT with softmax operation without
trainable parameters, i.e., weights of trees are determined in accordance with
(\ref{RF_Att_43}), and they are used to calculate the RF performance measures
as the weighted sum of the tree outcomes.

\item \textbf{ABRF-1}, \textbf{ABRF-2}, \textbf{ABRF-3}: one of the ABRF models.
\end{enumerate}

The best results in all tables are shown in bold. Moreover, for cases of
studying ABRF-1 and ABRF-3 models, the optimal values of the contamination
parameter $\epsilon_{opt}$ are provided. The case $\epsilon_{opt}=1$ means
that weights of trees are totally determined by the tree results and do not
depend on each instance. This case coincides with the weighted RF proposed in
\cite{Utkin-Konstantinov-etal-20}. The case $\epsilon_{opt}=0$ means that
weights of trees are determined only by the softmax function (with or without
trainable parameters).

All experiments can be divided into two groups:

\begin{enumerate}
\item The first group, called \textbf{Condition 1}, is based on training trees
in RFs or ERTs such that the largest depth of trees is 2. This condition is
used in order to ensure to have in each leaf more than one instance. At the
same time, the restriction of the tree depth may cause a lower quality of
trees. 

\item The second group, called \textbf{Condition 2}, is based on training
trees such that at least $10$ instances fall into every leaf of trees. This
condition is used to get desirable estimates of vectors $\mathbf{A}%
_{k}(\mathbf{x}_{s}\mathbf{)}$.
\end{enumerate}

Every RF or ERT consists of $100$ decision trees. To evaluate the average
accuracy measures, we perform a cross-validation with $100$ repetitions, where
in each run, we randomly select $n_{\text{tr}}=4n/5$ training data and
$n_{\text{test}}=n/5$ testing data.

\subsubsection{ABRF-1}

\begin{figure}
[ptb]
\begin{center}
\includegraphics[
height=2.9827in,
width=5.632in
]%
{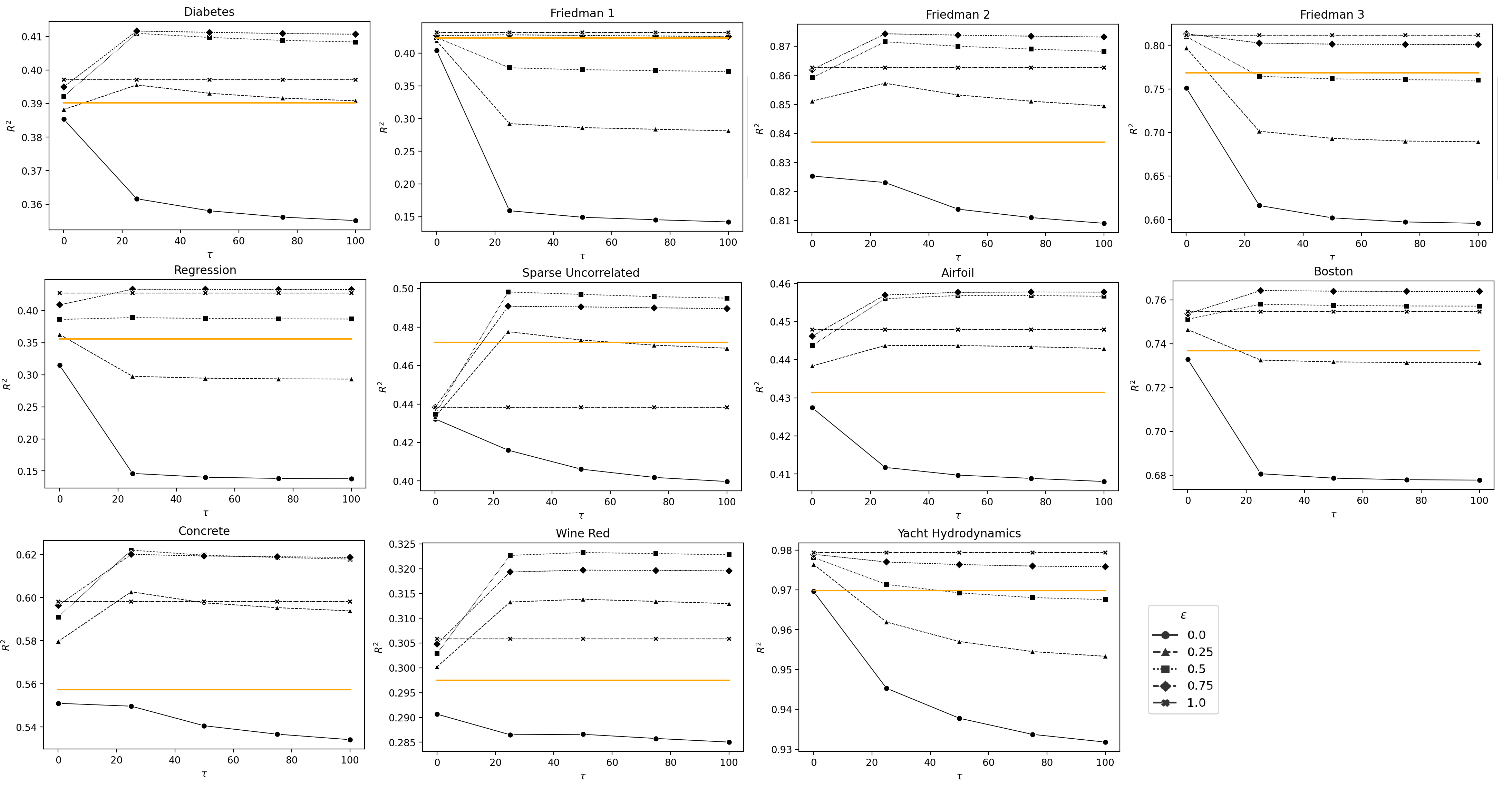}%
\caption{Measures $R^{2}$ as functions of the tuning parameter $\tau$ for $5$
different values of the tuning parameter $\epsilon$ computed for each dataset
}%
\label{f:tau_eps_depth_2}%
\end{center}
\end{figure}

It should be noted that ABRF-1 has two tuning parameters $\epsilon$ and $\tau
$, which may significantly impact on predictions. Therefore, the best
predictions are calculated at a predefined grid of the tuning parameters, and
a cross-validation procedure is subsequently used to select an appropriate
values of $\epsilon$ and $\tau$. Fig. \ref{f:tau_eps_depth_2} demonstrates how
parameters $\epsilon$ and $\tau$ impact on the performance measure ($R^{2}$)
for all considered datasets under Condition 1. Each picture corresponding to a
dataset in Fig. \ref{f:tau_eps_depth_2} shows measures $R^{2}$ as functions of
$\tau$ for $5$ different values of $\epsilon$ such that each function
corresponds to one value of $\epsilon$. In particular, curves with circle,
triangle, square, diamond, cross markers correspond to values $0$, $0.25$,
$0.5$, $0.75$, $1$ of $\epsilon$, respectively. The solid orange curve in each
picture corresponds to $R^{2}$ when the original RF without attention-based
model is used. The solid curve is depicted in order to illustrate the
relationship between the original RF and ABRF-1 by different values of
$\epsilon$ and $\tau$. It can be seen from Fig. \ref{f:tau_eps_depth_2} that
there is an optimal pair of values of $\epsilon$ and $\tau$ for each dataset,
which provides the largest value of $R^{2}$.%

Measures $R^{2}$ and MAE for three cases (RF, Softmax and ABRF-1) are shown in
Table \ref{t:regression_1_1}. The results are obtained by training the RF and
the parameter vector $\mathbf{w}$ on the regression datasets under Condition 1
that the largest depth of trees is 2. It can be seen from Table
\ref{t:regression_1_1} that ABRF-1 outperforms the RF and the Softmax models
almost for all datasets. The same results are shown in Table
\ref{t:regression_1_2} under Condition 2 that trees are built to ensure at
least $10$ instances in each leaf. One can again see from Table
\ref{t:regression_1_2} that ABRF-1 outperforms the RF and the Softmax models
for most datasets.

\begin{figure}
[ptb]
\begin{center}
\includegraphics[
height=2.308in,
width=3.5727in
]%
{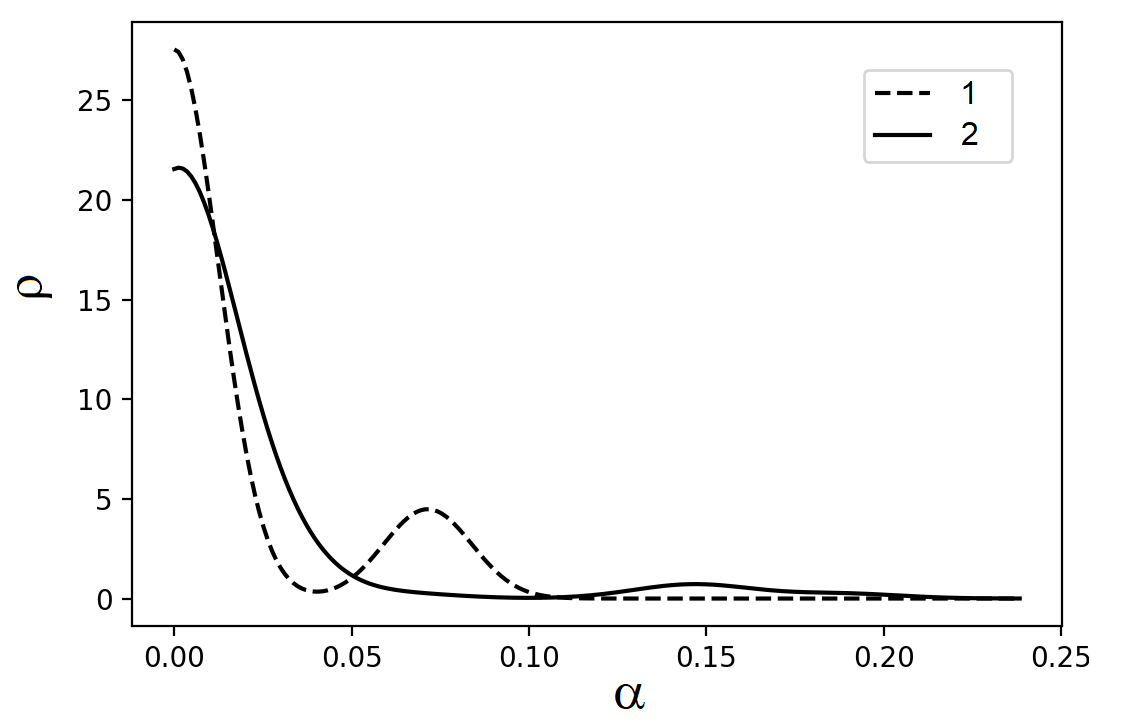}%
\caption{An example of the kernel density estimation $\rho$ as a function of
the attention weight distribution over trees}%
\label{f:kde}%
\end{center}
\end{figure}

An example of two kernel density estimations (KDEs) $\rho$ as functions of the
attention weight distributions over trees for a randomly selected instance is
depicted in Fig. \ref{f:kde}. The first KDE (the dash line 1) is constructed
using the attention weights computed by means of the softmax operation without
trainable parameters (the Softmax case). The second KDE (the solid line 2) is
constructed using the ABRF-1 model. The Gaussian kernel with the unit variance
is applied to constructing the KDEs. It can be seen from Fig. \ref{f:kde} that
the attention weight distribution computed by using ABRF-1 differs from the
first weight distribution. It can be regarded as a smooth version of the first
distribution.
%

\begin{table}[tbp] \centering
\caption{Measures $R^2$ and MAE for comparison of models (the RF, the Softmax model, ABRF-1) trained on regression datasets under Condition 1}%
\begin{tabular}
[c]{cccccccc}\hline
&  & \multicolumn{3}{c}{$R^{2}$} & \multicolumn{3}{c}{MAE}\\\hline
Data set & $\epsilon_{opt}$ & RF & Softmax & ABRF-1 & RF & Softmax &
ABRF-1\\\hline
Diabetes & $1$ & $0.390$ & $0.399$ & $\mathbf{0.405}$ & $\mathbf{46.06}$ &
$46.25$ & $46.15$\\\hline
Friedman 1 & $0.444$ & $0.423$ & $0.404$ & $\mathbf{0.432}$ & $2.624$ &
$2.655$ & $\mathbf{2.566}$\\\hline
Friedman 2 & $0.889$ & $0.837$ & $0.841$ & $\mathbf{0.868}$ & $113.9$ &
$113.6$ & $\mathbf{106.5}$\\\hline
Friedman 3 & $0.889$ & $0.769$ & $0.752$ & $\mathbf{0.812}$ & $0.133$ &
$0.138$ & $\mathbf{0.124}$\\\hline
Regression & $0.333$ & $0.356$ & $0.322$ & $\mathbf{0.431}$ & $111.2$ &
$114.6$ & $\mathbf{102.2}$\\\hline
Sparse & $0$ & $0.472$ & $\mathbf{0.534}$ & $\mathbf{0.534}$ & $1.915$ &
$\mathbf{1.766}$ & $\mathbf{1.766}$\\\hline
Airfoil & $1$ & $0.431$ & $0.427$ & $\mathbf{0.448}$ & $4.157$ & $4.175$ &
$\mathbf{4.082}$\\\hline
Boston & $0.556$ & $0.737$ & $0.755$ & $\mathbf{0.765}$ & $3.266$ & $3.175$ &
$\mathbf{3.123}$\\\hline
Concrete & $0.556$ & $0.557$ & $0.571$ & $\mathbf{0.607}$ & $8.680$ & $8.549$
& $\mathbf{8.201}$\\\hline
Wine & $0.444$ & $0.297$ & $0.304$ & $\mathbf{0.314}$ & $0.527$ & $0.527$ &
$\mathbf{0.524}$\\\hline
Yacht & $1$ & $0.970$ & $0.970$ & $\mathbf{0.979}$ & $1.911$ & $1.920$ &
$\mathbf{1.540}$\\\hline
\end{tabular}
\label{t:regression_1_1}%
\end{table}%
%

\begin{table}[tbp] \centering
\caption{Measures $R^2$ and MAE for comparison of models (the RF, the Softmax model, ABRF-1) trained on regression datasets under Condition 2}%
\begin{tabular}
[c]{cccccccc}\hline
&  & \multicolumn{3}{c}{$R^{2}$} & \multicolumn{3}{c}{MAE}\\\hline
Data set & $\epsilon$ & RF & Softmax & ABRF-1 & RF & Softmax & ABRF-1\\\hline
Diabetes & $0$ & $0.416$ & $\mathbf{0.424}$ & $\mathbf{0.424}$ & $44.92$ &
$\mathbf{44.66}$ & $\mathbf{44.66}$\\\hline
Friedman 1 & $1$ & $0.459$ & $0.438$ & $\mathbf{0.470}$ & $\mathbf{2.540}$ &
$2.589$ & $\mathbf{2.540}$\\\hline
Friedman 2 & $0.556$ & $0.841$ & $0.847$ & $\mathbf{0.877}$ & $111.7$ &
$110.0$ & $\mathbf{102.0}$\\\hline
Friedman 3 & $0.667$ & $0.625$ & $0.628$ & $\mathbf{0.686}$ & $0.154$ &
$0.155$ & $\mathbf{0.134}$\\\hline
Regression & $0.667$ & $0.380$ & $0.366$ & $\mathbf{0.450}$ & $109.1$ &
$110.5$ & $\mathbf{100.8}$\\\hline
Sparse & $0$ & $0.470$ & $\mathbf{0.529}$ & $\mathbf{0.529}$ & $1.908$ &
$1.790$ & $\mathbf{1.790}$\\\hline
Airfoil & $1$ & $0.823$ & $0.820$ & $\mathbf{0.843}$ & $2.203$ & $2.231$ &
$\mathbf{2.070}$\\\hline
Boston & $0.556$ & $0.814$ & $0.819$ & $\mathbf{0.823}$ & $2.539$ & $2.508$ &
$\mathbf{2.494}$\\\hline
Concrete & $1$ & $0.845$ & $0.841$ & $\mathbf{0.857}$ & $4.855$ & $4.948$ &
$\mathbf{4.694}$\\\hline
Wine & $0.667$ & $\mathbf{0.433}$ & $0.422$ & $0.423$ & $\mathbf{0.451}$ &
$0.461$ & $0.459$\\\hline
Yacht & $1$ & $0.981$ & $0.981$ & $\mathbf{0.989}$ & $1.004$ & $1.004$ &
$\mathbf{0.787}$\\\hline
\end{tabular}
\label{t:regression_1_2}%
\end{table}%

Another interesting question is how the attention-based model performs when
ERTs are used. The corresponding results under Condition 1 and Condition 2 are
shown in Tables \ref{t:regression_1_3} and \ref{t:regression_1_4},
respectively. It is interesting to point out that ERTs provide comparable and
even better results than RFs.%

\begin{table}[tbp] \centering
\caption{Measures $R^2$ and MAE for comparison of models (the ERT, the Softmax model, ABRF-1) trained on regression datasets under Condition 1}%
\begin{tabular}
[c]{cccccccc}\hline
&  & \multicolumn{3}{c}{$R^{2}$} & \multicolumn{3}{c}{MAE}\\\hline
Data set & $\epsilon_{opt}$ & ERT & Softmax & ABRF-1 & ERT & Softmax &
ABRF-1\\\hline
Diabetes & $0.333$ & $0.376$ & $0.400$ & $\mathbf{0.407}$ & $48.439$ &
$47.382$ & $\mathbf{46.721}$\\\hline
Friedman 1 & $0.778$ & $0.411$ & $0.416$ & $\mathbf{0.449}$ & $2.638$ &
$2.639$ & $\mathbf{2.533}$\\\hline
Friedman 2 & $0.556$ & $0.782$ & $0.810$ & $\mathbf{0.872}$ & $142.2$ &
$131.8$ & $\mathbf{105.3}$\\\hline
Friedman 3 & $1$ & $0.606$ & $0.606$ & $\mathbf{0.760}$ & $0.178$ & $0.178$ &
$\mathbf{0.137}$\\\hline
Regression & $1$ & $0.352$ & $0.352$ & $\mathbf{0.451}$ & $110.7$ & $110.7$ &
$\mathbf{100.5}$\\\hline
Sparse & $0$ & $0.394$ & $\mathbf{0.489}$ & $\mathbf{0.489}$ & $2.097$ &
$\mathbf{1.888}$ & $\mathbf{1.888}$\\\hline
Airfoil & $0.667$ & $0.305$ & $0.342$ & $\mathbf{0.447}$ & $4.656$ & $4.545$ &
$\mathbf{4.106}$\\\hline
Boston & $0.667$ & $0.690$ & $0.721$ & $\mathbf{0.769}$ & $3.573$ & $3.412$ &
$\mathbf{3.097}$\\\hline
Concrete & $0.778$ & $0.416$ & $0.433$ & $\mathbf{0.555}$ & $10.14$ & $10.02$
& $\mathbf{8.77}$\\\hline
Wine & $0.333$ & $0.257$ & $0.298$ & $\mathbf{0.315}$ & $0.550$ & $0.532$ &
$\mathbf{0.524}$\\\hline
Yacht & $1$ & $0.883$ & $0.883$ & $\mathbf{0.985}$ & $3.218$ & $3.218$ &
$\mathbf{1.105}$\\\hline
\end{tabular}
\label{t:regression_1_3}%
\end{table}%
%

\begin{table}[tbp] \centering
\caption{Measures $R^2$ and MAE for comparison of models (the ERT, the Softmax model, ABRF-1) trained on regression datasets under Condition 2}%
\begin{tabular}
[c]{cccccccc}\hline
&  & \multicolumn{3}{c}{$R^{2}$} & \multicolumn{3}{c}{MAE}\\\hline
Data set & $\epsilon$ & ERT & Softmax & ABRF-1 & ERT & Softmax &
ABRF-1\\\hline
Diabetes & $0.222$ & $0.438$ & $\mathbf{0.441}$ & $\mathbf{0.441}$ & $44.31$ &
$44.00$ & $\mathbf{43.98}$\\\hline
Friedman 1 & $1$ & $0.471$ & $0.471$ & $\mathbf{0.513}$ & $2.502$ & $2.502$ &
$\mathbf{2.426}$\\\hline
Friedman 2 & $1$ & $0.813$ & $0.813$ & $\mathbf{0.930}$ & $123.0$ & $123.0$ &
$\mathbf{74.596}$\\\hline
Friedman 3 & $1$ & $0.570$ & $0.570$ & $\mathbf{0.739}$ & $0.179$ & $0.179$ &
$\mathbf{0.138}$\\\hline
Regression & $0.889$ & $0.402$ & $0.405$ & $\mathbf{0.447}$ & $106.3$ &
$106.1$ & $\mathbf{102.1}$\\\hline
Sparse & $0.111$ & $0.452$ & $0.527$ & $\mathbf{0.536}$ & $1.994$ & $1.845$ &
$\mathbf{1.826}$\\\hline
Airfoil & $1$ & $0.802$ & $0.802$ & $\mathbf{0.837}$ & $2.370$ & $2.370$ &
$\mathbf{2.128}$\\\hline
Boston & $0.889$ & $0.831$ & $0.833$ & $\mathbf{0.838}$ & $2.481$ & $2.468$ &
$\mathbf{2.454}$\\\hline
Concrete & $1$ & $0.851$ & $0.851$ & $\mathbf{0.863}$ & $4.892$ & $4.892$ &
$\mathbf{4.650}$\\\hline
Wine & $0.889$ & $\mathbf{0.418}$ & $0.417$ & $0.416$ & $0.464$ &
$\mathbf{0.463}$ & $\mathbf{0.463}$\\\hline
Yacht & $1$ & $\mathbf{0.988}$ & $\mathbf{0.988}$ & $\mathbf{0.988}$ & $0.824$
& $0.824$ & $\mathbf{0.818}$\\\hline
\end{tabular}
\label{t:regression_1_4}%
\end{table}%

Performance measures of the linear modification of ABRF-1, when it is trained
by using the linear optimization problem (\ref{RF_Att_34})-(\ref{RF_Att_36})
under Condition 2 for the RF and for the ERT, are given in Tables
\ref{t:regression-1-lin-2} and \ref{t:regression-1-lin-1}, respectively. To
compare the performance results obtained by means of the linear and quadratic
optimization problems, we collect values of the $R^{2}$ measure for ABRF-1
from Table \ref{t:regression-1-lin-2} and from Table \ref{t:regression_1_2} in
Table \ref{t:regression_1_l_q}. It can be seen from Table
\ref{t:regression_1_l_q} that the quadratic optimization problem provides
better results in comparison with the linear case. However, this
outperformance is rather negligible. The same can be said if to compare the
results under other conditions (ERTs and Condition 1).%

\begin{table}[tbp] \centering
\caption{Measures $R^2$ and MAE for comparison of models (the RF, the Softmax model, ABRF-1) trained on regression datasets under Condition 2 and the $L_1$-norm is used for loss function (the linear programming case)}%
\begin{tabular}
[c]{cccccccc}\hline
&  & \multicolumn{3}{c}{$R^{2}$} & \multicolumn{3}{c}{MAE}\\\hline
Data set & $\epsilon$ & RF & Softmax & ABRF-1 & RF & Softmax & ABRF-1\\\hline
Diabetes & $0$ & $0.416$ & $\mathbf{0.424}$ & $\mathbf{0.424}$ & $44.92$ &
$\mathbf{44.66}$ & $\mathbf{44.66}$\\\hline
Friedman 1 & $0.667$ & $\mathbf{0.459}$ & $0.433$ & $0.454$ & $\mathbf{2.540}$
& $2.589$ & $2.594$\\\hline
Friedman 2 & $0.556$ & $0.841$ & $0.847$ & $\mathbf{0.877}$ & $111.7$ &
$110.0$ & $\mathbf{101.7}$\\\hline
Friedman 3 & $0.667$ & $0.625$ & $0.628$ & $\mathbf{0.689}$ & $0.154$ &
$0.155$ & $\mathbf{0.132}$\\\hline
Regression & $0.556$ & $0.380$ & $0.366$ & $\mathbf{0.439}$ & $109.1$ &
$110.5$ & $\mathbf{100.7}$\\\hline
Sparse & $0$ & $0.470$ & $\mathbf{0.529}$ & $\mathbf{0.529}$ & $1.908$ &
$\mathbf{1.790}$ & $\mathbf{1.790}$\\\hline
Airfoil & $1$ & $0.823$ & $0.820$ & $\mathbf{0.843}$ & $2.203$ & $2.231$ &
$\mathbf{2.064}$\\\hline
Boston & $0.444$ & $0.814$ & $0.819$ & $\mathbf{0.824}$ & $2.539$ & $2.508$ &
$\mathbf{2.468}$\\\hline
Concrete & $1$ & $0.845$ & $0.841$ & $\mathbf{0.857}$ & $4.834$ & $4.948$ &
$\mathbf{4.622}$\\\hline
Wine & $0.222$ & $\mathbf{0.433}$ & $0.422$ & $0.418$ & $\mathbf{0.451}$ &
$0.461$ & $0.460$\\\hline
Yacht & $1$ & $0.981$ & $0.981$ & $\mathbf{0.988}$ & $1.004$ & $1.004$ &
$\mathbf{0.808}$\\\hline
\end{tabular}
\label{t:regression-1-lin-2}%
\end{table}%
%

\begin{table}[tbp] \centering
\caption{Measures $R^2$ and MAE for comparison of models (the ERT, the Softmax model, ABRF-1) trained on regression datasets under Condition 2 and the $L_1$-norm is used for loss function (the linear programming case)}%
\begin{tabular}
[c]{cccccccc}\hline
&  & \multicolumn{3}{c}{$R^{2}$} & \multicolumn{3}{c}{MAE}\\\hline
Data set & $\epsilon$ & ERT & Softmax & ABRF-1 & ERT & Softmax &
ABRF-1\\\hline
Diabetes & $0$ & $\mathbf{0.438}$ & $0.431$ & $0.433$ & $44.55$ &
$\mathbf{44.00}$ & $44.43$\\\hline
Friedman 1 & $0.667$ & $0.471$ & $0.471$ & $\mathbf{0.498}$ & $2.502$ &
$2.502$ & $\mathbf{2.453}$\\\hline
Friedman 2 & $1$ & $0.813$ & $0.813$ & $\mathbf{0.924}$ & $123.0$ & $123.0$ &
$\mathbf{76.34}$\\\hline
Friedman 3 & $0.667$ & $0.570$ & $0.570$ & $\mathbf{0.699}$ & $0.179$ &
$0.179$ & $\mathbf{0.143}$\\\hline
Regression & $0.889$ & $0.402$ & $0.405$ & $\mathbf{0.445}$ & $106.3$ &
$106.1$ & $\mathbf{102.7}$\\\hline
Sparse & $0$ & $0.452$ & $0.527$ & $\mathbf{0.534}$ & $1.994$ & $1.845$ &
$\mathbf{1.830}$\\\hline
Airfoil & $1$ & $0.802$ & $0.802$ & $\mathbf{0.835}$ & $2.370$ & $2.370$ &
$\mathbf{2.130}$\\\hline
Boston & $0.889$ & $0.831$ & $0.833$ & $\mathbf{0.839}$ & $2.481$ & $2.468$ &
$\mathbf{2.396}$\\\hline
Concrete & $1$ & $0.839$ & $0.851$ & $\mathbf{0.856}$ & $5.119$ & $4.892$ &
$\mathbf{4.721}$\\\hline
Wine & $0.556$ & $\mathbf{0.418}$ & $0.417$ & $0.411$ & $0.464$ & $0.463$ &
$\mathbf{0.461}$\\\hline
Yacht & $1$ & $\mathbf{0.988}$ & $\mathbf{0.988}$ & $\mathbf{0.988}$ & $0.824$
& $0.824$ & $\mathbf{0.818}$\\\hline
\end{tabular}
\label{t:regression-1-lin-1}%
\end{table}%
%

\begin{table}[tbp] \centering
\caption{Measure $R^2$ for comparison of ABRF-1 trained on regression datasets with using the linear and the quadratic optimization problems}%
\begin{tabular}
[c]{ccc}\hline
Data set & Linear & Quadratic\\\hline
Diabetes & $0.424$ & $0.424$\\\hline
Friedman 1 & $0.454$ & $\mathbf{0.470}$\\\hline
Friedman 2 & $0.877$ & $0.877$\\\hline
Friedman 3 & $\mathbf{0.689}$ & $0.686$\\\hline
Regression & $0.439$ & $\mathbf{0.450}$\\\hline
Sparse & $0.529$ & $0.529$\\\hline
Airfoil & $0.843$ & $0.843$\\\hline
Boston & $\mathbf{0.824}$ & $0.823$\\\hline
Concrete & $0.857$ & $0.857$\\\hline
Wine & $0.418$ & $\mathbf{0.423}$\\\hline
Yacht & $0.988$ & $\mathbf{0.989}$\\\hline
\end{tabular}
\label{t:regression_1_l_q}%
\end{table}%

\subsubsection{ABRF-2}

Numerical experiments with ABRF-2 are presented in Tables
\ref{t:regression_ABRF-2}-\ref{t:regression_ABRF-2-3}. Since ABRF-2 is based
only on the trainable softmax operation and does not use the contamination
model, then we can take $\epsilon=0$ in all experiments with ABRF-2, and
values of $\epsilon$ are not presented in tables with results. It can be seen
from Tables \ref{t:regression_ABRF-2}-\ref{t:regression_ABRF-2-3} that ABRF-2
does not outperform the RF (the ERT) for many datasets. This implies that the
attention-based model without the contamination component provides worse
results in comparison with ABRF-1.%

\begin{table}[tbp] \centering
\caption{Measures $R^2$ and MAE for comparison of models (the RF, the Softmax model, ABRF-2) trained on regression datasets under Condition 1}%
\begin{tabular}
[c]{ccccccc}\hline
& \multicolumn{3}{c}{$R^{2}$} & \multicolumn{3}{c}{MAE}\\\hline
Data set & RF & Softmax & ABRF-2 & RF & Softmax & ABRF-2\\\hline
Diabetes & $0.390$ & $0.399$ & $\mathbf{0.408}$ & $46.06$ & $46.25$ &
$\mathbf{45.93}$\\\hline
Friedman 1 & $\mathbf{0.423}$ & $0.404$ & $0.303$ & $\mathbf{2.624}$ & $2.655$
& $2.775$\\\hline
Friedman 2 & $0.837$ & $0.841$ & $\mathbf{0.933}$ & $113.9$ & $113.6$ &
$\mathbf{64.66}$\\\hline
Friedman 3 & $\mathbf{0.769}$ & $0.752$ & $0.621$ & $\mathbf{0.133}$ & $0.138$
& $0.158$\\\hline
Regression & $\mathbf{0.356}$ & $0.322$ & $0.336$ & $\mathbf{111.2}$ & $114.6$
& $111.7$\\\hline
Sparse & $0.472$ & $\mathbf{0.534}$ & $0.490$ & $1.915$ & $1.766$ &
$\mathbf{1.746}$\\\hline
Airfoil & $0.431$ & $0.427$ & $\mathbf{0.451}$ & $4.157$ & $4.175$ &
$\mathbf{4.047}$\\\hline
Boston & $0.737$ & $0.755$ & $\mathbf{0.783}$ & $3.266$ & $3.175$ &
$\mathbf{2.976}$\\\hline
Concrete & $0.557$ & $0.571$ & $\mathbf{0.666}$ & $8.680$ & $8.549$ &
$\mathbf{7.524}$\\\hline
Wine & $0.297$ & $0.304$ & $\mathbf{0.345}$ & $0.527$ & $0.527$ &
$\mathbf{0.500}$\\\hline
Yacht & $0.970$ & $0.970$ & $\mathbf{0.971}$ & $1.911$ & $1.920$ &
$\mathbf{1.700}$\\\hline
\end{tabular}
\label{t:regression_ABRF-2}%
\end{table}%
%

\begin{table}[tbp] \centering
\caption{Measures $R^2$ and MAE for comparison of models (the ERT, the Softmax model,  ABRF-2) trained on regression datasets under Condition 2}%
\begin{tabular}
[c]{ccccccc}\hline
& \multicolumn{3}{c}{$R^{2}$} & \multicolumn{3}{c}{MAE}\\\hline
Data set & ERT & Softmax & ABRF-2 & ERT & Softmax & ABRF-2\\\hline
Diabetes & $0.438$ & $\mathbf{0.441}$ & $0.345$ & $44.55$ & $\mathbf{44.00}$ &
$47.57$\\\hline
Friedman 1 & $\mathbf{0.471}$ & $\mathbf{0.471}$ & $0.281$ & $\mathbf{2.502}$
& $\mathbf{2.502}$ & $2.775$\\\hline
Friedman 2 & $0.813$ & $0.813$ & $\mathbf{0.945}$ & $123.0$ & $123.0$ &
$\mathbf{50.88}$\\\hline
Friedman 3 & $0.570$ & $0.570$ & $\mathbf{0.662}$ & $0.179$ & $0.179$ &
$\mathbf{0.156}$\\\hline
Regression & $0.402$ & $0.405$ & $\mathbf{0.516}$ & $106.3$ & $106.1$ &
$\mathbf{91.58}$\\\hline
Sparse & $0.452$ & $0.527$ & $\mathbf{0.591}$ & $1.994$ & $1.845$ &
$\mathbf{1.582}$\\\hline
Airfoil & $\mathbf{0.802}$ & $\mathbf{0.802}$ & $0.772$ & $\mathbf{2.370}$ &
$\mathbf{2.370}$ & $2.471$\\\hline
Boston & $0.831$ & $\mathbf{0.833}$ & $0.802$ & $2.481$ & $\mathbf{2.468}$ &
$2.704$\\\hline
Concrete & $0.839$ & $\mathbf{0.851}$ & $0.743$ & $5.119$ & $\mathbf{4.892}$ &
$6.126$\\\hline
Wine & $\mathbf{0.418}$ & $0.417$ & $0.312$ & $0.464$ & $\mathbf{0.463}$ &
$0.495$\\\hline
Yacht & $\mathbf{0.988}$ & $\mathbf{0.988}$ & $\mathbf{0.988}$ & $0.824$ &
$0.824$ & $\mathbf{0.818}$\\\hline
\end{tabular}
\label{t:regression_ABRF-2-2}%
\end{table}%
%

\begin{table}[tbp] \centering
\caption{Measures $R^2$ and MAE for comparison of models (the RF, theSoftmax model, ABRF-2) trained on regression datasets under Condition 2}%
\begin{tabular}
[c]{ccccccc}\hline
& \multicolumn{3}{c}{$R^{2}$} & \multicolumn{3}{c}{MAE}\\\hline
Data set & RF & Softmax & ABRF-2 & RF & Softmax & ABRF-2\\\hline
Diabetes & $0.416$ & $\mathbf{0.424}$ & $0.328$ & $44.92$ & $\mathbf{44.66}$ &
$47.59$\\\hline
Friedman 1 & $\mathbf{0.459}$ & $0.438$ & $0.325$ & $\mathbf{2.540}$ & $2.589$
& $2.792$\\\hline
Friedman 2 & $0.841$ & $0.847$ & $\mathbf{0.929}$ & $111.7$ & $110.0$ &
$\mathbf{69.1}$\\\hline
Friedman 3 & $0.625$ & $0.628$ & $\mathbf{0.641}$ & $0.154$ & $0.155$ &
$\mathbf{0.145}$\\\hline
Regression & $0.380$ & $0.366$ & $\mathbf{0.570}$ & $109.1$ & $110.5$ &
$\mathbf{88.21}$\\\hline
Sparse & $0.470$ & $0.529$ & $\mathbf{0.532}$ & $1.908$ & $1.790$ &
$\mathbf{1.671}$\\\hline
Airfoil & $\mathbf{0.823}$ & $0.820$ & $0.768$ & $\mathbf{2.203}$ & $2.231$ &
$2.501$\\\hline
Boston & $0.814$ & $\mathbf{0.819}$ & $0.779$ & $2.539$ & $\mathbf{2.508}$ &
$2.750$\\\hline
Concrete & $\mathbf{0.845}$ & $0.841$ & $0.786$ & $\mathbf{4.834}$ & $4.948$ &
$5.591$\\\hline
Wine & $\mathbf{0.433}$ & $0.422$ & $0.354$ & $\mathbf{0.451}$ & $0.461$ &
$0.481$\\\hline
Yacht & $\mathbf{0.981}$ & $\mathbf{0.981}$ & $\mathbf{0.981}$ &
$\mathbf{1.004}$ & $\mathbf{1.004}$ & $1.057$\\\hline
\end{tabular}
\label{t:regression_ABRF-2-3}%
\end{table}%

\subsubsection{ABRF-3}

Numerical experiments with ABRF-2 are presented in Tables
\ref{t:regression_ABRF-3-1}-\ref{t:regression_ABRF-3-3}. One can see from the
tables that ABRF-3 outperforms other models. In particular, as it is shown in
Table \ref{t:regression_ABRF-3-1}, ABRF-3 provides better results for all
datasets under Condition 1. In fact, this model clearly corrects inaccurate
predictions of RFs trained under Condition 1. This is a very important
property of ABRF-3. It should be noted that similar results were demonstrated
by ABRF-1 when Condition 1 was used to train trees. However, it follows, for
example, from Table \ref{t:regression_1_1}, that there are a few datasets for
which ABRF-1 provides worse results.%

\begin{table}[tbp] \centering
\caption{Measures $R^2$ and MAE for comparison of models (the RF, the Softmax model, ABRF-3) trained on regression datasets under Condition 1}%
\begin{tabular}
[c]{cccccccc}\hline
&  & \multicolumn{3}{c}{$R^{2}$} & \multicolumn{3}{c}{MAE}\\\hline
Data set & $\epsilon$ & RF & Softmax & ABRF-3 & RF & Softmax & ABRF-3\\\hline
Diabetes & $0$ & $0.390$ & $0.399$ & $\mathbf{0.408}$ & $46.06$ & $46.25$ &
$\mathbf{45.93}$\\\hline
Friedman 1 & $0.778$ & $0.423$ & $0.404$ & $\mathbf{0.428}$ & $2.624$ &
$2.655$ & $\mathbf{2.605}$\\\hline
Friedman 2 & $0$ & $0.837$ & $0.841$ & $\mathbf{0.933}$ & $113.9$ & $113.6$ &
$\mathbf{64.66}$\\\hline
Friedman 3 & $1$ & $0.769$ & $0.752$ & $\mathbf{0.811}$ & $0.133$ & $0.138$ &
$\mathbf{0.124}$\\\hline
Regression & $1$ & $0.356$ & $0.322$ & $\mathbf{0.424}$ & $111.2$ & $114.6$ &
$\mathbf{102.9}$\\\hline
Sparse & $0.111$ & $0.472$ & $0.534$ & $\mathbf{0.602}$ & $1.915$ & $1.766$ &
$\mathbf{1.596}$\\\hline
Airfoil & $0.222$ & $0.431$ & $0.427$ & $\mathbf{0.457}$ & $4.157$ & $4.175$ &
$\mathbf{4.023}$\\\hline
Boston & $0.222$ & $0.737$ & $0.755$ & $\mathbf{0.788}$ & $3.266$ & $3.175$ &
$\mathbf{2.975}$\\\hline
Concrete & $0$ & $0.557$ & $0.571$ & $\mathbf{0.666}$ & $8.680$ & $8.549$ &
$\mathbf{7.524}$\\\hline
Wine & $0$ & $0.297$ & $0.304$ & $\mathbf{0.345}$ & $0.527$ & $0.527$ &
$\mathbf{0.500}$\\\hline
Yacht & $0.889$ & $0.970$ & $0.970$ & $\mathbf{0.980}$ & $1.911$ & $1.920$ &
$\mathbf{1.525}$\\\hline
\end{tabular}
\label{t:regression_ABRF-3-1}%
\end{table}%
%

\begin{table}[tbp] \centering
\caption{Measures $R^2$ and MAE for comparison of models (the ERT, the Softmax model, ABRF-3) trained on regression datasets under Condition 2}%
\begin{tabular}
[c]{cccccccc}\hline
&  & \multicolumn{3}{c}{$R^{2}$} & \multicolumn{3}{c}{MAE}\\\hline
Data set & $\epsilon$ & ERT & Softmax & ABRF-3 & ERT & Softmax &
ABRF-3\\\hline
Diabetes & $1$ & $\mathbf{0.438}$ & $\mathbf{0.438}$ & $0.425$ & $44.55$ &
$\mathbf{44.00}$ & $44.90$\\\hline
Friedman 1 & $1$ & $0.471$ & $0.471$ & $\mathbf{0.505}$ & $2.502$ & $2.502$ &
$\mathbf{2.443}$\\\hline
Friedman 2 & $0.111$ & $0.813$ & $0.836$ & $\mathbf{0.966}$ & $123.0$ &
$123.0$ & $\mathbf{44.651}$\\\hline
Friedman 3 & $0.889$ & $0.570$ & $0.570$ & $\mathbf{0.709}$ & $0.179$ &
$0.179$ & $\mathbf{0.140}$\\\hline
Regression & $0.333$ & $0.402$ & $0.384$ & $\mathbf{0.518}$ & $106.3$ &
$106.1$ & $\mathbf{92.720}$\\\hline
Sparse & $0.222$ & $0.452$ & $0.520$ & $\mathbf{0.629}$ & $1.994$ & $1.845$ &
$\mathbf{1.556}$\\\hline
Airfoil & $1$ & $0.802$ & $0.802$ & $\mathbf{0.836}$ & $2.370$ & $2.370$ &
$\mathbf{2.129}$\\\hline
Boston & $1$ & $0.831$ & $0.831$ & $\mathbf{0.836}$ & $2.481$ & $2.468$ &
$\mathbf{2.465}$\\\hline
Concrete & $1$ & $0.839$ & $0.839$ & $\mathbf{0.857}$ & $5.119$ & $4.892$ &
$\mathbf{4.776}$\\\hline
Wine & $1$ & $\mathbf{0.418}$ & $\mathbf{0.418}$ & $0.415$ & $0.464$ & $0.463$
& $\mathbf{0.462}$\\\hline
Yacht & $0.444$ & $\mathbf{0.988}$ & $\mathbf{0.988}$ & $\mathbf{0.988}$ &
$0.824$ & $0.824$ & $\mathbf{0.817}$\\\hline
\end{tabular}
\label{t:regression_ABRF-3-2}%
\end{table}%
%

\begin{table}[tbp] \centering
\caption{Measures $R^2$ and MAE for comparison of models (the RF, the Softmax model, ABRF-3) trained on regression datasets under Condition 2}%
\begin{tabular}
[c]{cccccccc}\hline
&  & \multicolumn{3}{c}{$R^{2}$} & \multicolumn{3}{c}{MAE}\\\hline
Data set & $\epsilon$ & RF & Softmax & ABRF-3 & RF & Softmax & ABRF-3\\\hline
Diabetes & $0.778$ & $0.416$ & $\mathbf{0.424}$ & $0.406$ & $44.92$ &
$\mathbf{44.66}$ & $45.36$\\\hline
Friedman 1 & $1$ & $\mathbf{0.459}$ & $0.438$ & $\mathbf{0.459}$ &
$\mathbf{2.540}$ & $2.589$ & $2.564$\\\hline
Friedman 2 & $0.222$ & $0.841$ & $0.847$ & $\mathbf{0.933}$ & $111.7$ &
$110.0$ & $\mathbf{68.93}$\\\hline
Friedman 3 & $0.111$ & $0.625$ & $0.628$ & $\mathbf{0.688}$ & $0.154$ &
$0.155$ & $\mathbf{0.140}$\\\hline
Regression & $0.111$ & $0.380$ & $0.366$ & $\mathbf{0.578}$ & $109.1$ &
$110.5$ & $\mathbf{87.87}$\\\hline
Sparse & $0.111$ & $0.470$ & $0.529$ & $\mathbf{0.592}$ & $1.908$ & $1.790$ &
$\mathbf{1.599}$\\\hline
Airfoil & $0.889$ & $0.823$ & $0.820$ & $\mathbf{0.844}$ & $2.203$ & $2.231$ &
$\mathbf{2.066}$\\\hline
Boston & $0.222$ & $0.814$ & $0.819$ & $\mathbf{0.823}$ & $2.539$ &
$\mathbf{2.508}$ & $2.520$\\\hline
Concrete & $1$ & $0.845$ & $0.841$ & $\mathbf{0.857}$ & $4.834$ & $4.948$ &
$\mathbf{4.677}$\\\hline
Wine & $1$ & $\mathbf{0.433}$ & $0.422$ & $0.421$ & $\mathbf{0.451}$ & $0.461$
& $0.459$\\\hline
Yacht & $1$ & $0.981$ & $0.981$ & $\mathbf{0.989}$ & $1.004$ & $1.004$ &
$\mathbf{0.789}$\\\hline
\end{tabular}
\label{t:regression_ABRF-3-3}%
\end{table}%

In Table \ref{t:regression_ABRF-123}, we compare all proposed attention-based
models (ABRF-1, ABRF-2, ABRF-3) by using the measure $R^{2}$ and the same
conditions (RFs built to ensure at least $10$ instances in every leaf). It can
be seen from Table \ref{t:regression_ABRF-123} that ABRF-3 is comparable with
ABRF-1, but both the models outperform ABRF-2.%

\begin{table}[tbp] \centering
\caption{Measure $R^2$ for comparison of models (ABRF-1, ABRF-2, ABRF-3) trained on regression datasets under Condition 2}%
\begin{tabular}
[c]{cccc}\hline
Data set & ABRF-1 & ABRF-2 & ABRF-3\\\hline
Diabetes & $\mathbf{0.424}$ & $0.328$ & $0.406$\\\hline
Friedman 1 & $\mathbf{0.470}$ & $0.325$ & $0.459$\\\hline
Friedman 2 & $0.877$ & $0.929$ & $\mathbf{0.933}$\\\hline
Friedman 3 & $0.686$ & $0.641$ & $\mathbf{0.688}$\\\hline
Regression & $0.450$ & $0.570$ & $\mathbf{0.578}$\\\hline
Sparse & $0.529$ & $0.532$ & $\mathbf{0.592}$\\\hline
Airfoil & $0.843$ & $0.768$ & $\mathbf{0.844}$\\\hline
Boston & $\mathbf{0.823}$ & $0.779$ & $\mathbf{0.823}$\\\hline
Concrete & $\mathbf{0.857}$ & $0.786$ & $\mathbf{0.857}$\\\hline
Wine & $\mathbf{0.423}$ & $0.354$ & $0.421$\\\hline
Yacht & $\mathbf{0.989}$ & $0.981$ & $\mathbf{0.989}$\\\hline
\end{tabular}
\label{t:regression_ABRF-123}%
\end{table}%

\subsection{Classification}

To study the proposed attention-based models for solving classification
problems, we apply datasets which are taken from the UCI Machine Learning
Repository \cite{Dua:2019}, in particular, Diabetic Retinopathy, Eeg Eyes,
Haberman's Survival, Ionosphere, Seeds, Seismic-Bumps, Soybean, Teaching
Assistant Evaluation, Tic-Tac-Toe Endgame, Website Phishing, Wholesale
Customer. Table \ref{t:class_datasets} shows the number of features $m$ for
the corresponding data set, the number of instances $n$, the number of classes
$C$ and their abbreviation. More detailed information can be found from the
data resources.

$F_{1}$ score is used in the classification experiments as an accuracy measure
which takes into account the class imbalance. Every RF consists of $100$
decision trees. To evaluate the average accuracy measures, we again perform a
cross-validation with $100$ repetitions, where in each run, we randomly select
$n_{\text{tr}}=4n/5$ training data and $n_{\text{test}}=n/5$ testing data.%

\begin{table}[tbp] \centering
\caption{A brief introduction about the classification data sets}%
\begin{tabular}
[c]{ccccc}\hline
Data set & Abbreviation & $m$ & $n$ & $C$\\\hline
Diabetic Retinopathy & Diabet & $20$ & $1151$ & $2$\\\hline
Eeg Eyes & Eeg & $14$ & $14980$ & $2$\\\hline
Haberman's Survival & Haberman & $3$ & $306$ & $2$\\\hline
Ionosphere & Ionosphere & $34$ & $351$ & $2$\\\hline
Seeds & Seeds & $7$ & $210$ & $3$\\\hline
Seismic-Bumps & Seismic & $18$ & $2584$ & $2$\\\hline
Soybean & Soybean & $35$ & $47$ & $4$\\\hline
Teaching Assistant Evaluation & TAE & $5$ & $151$ & $3$\\\hline
Tic-Tac-Toe Endgame & TTTE & $27$ & $957$ & $2$\\\hline
Website Phishing & Phishing & $9$ & $1353$ & $3$\\\hline
Wholesale Customer & Wholesale & $6$ & $440$ & $3$\\\hline
\end{tabular}
\label{t:class_datasets}%
\end{table}%

Table \ref{t:classif_ABRF-1-1} illustrates results of numerical experiments in
the form of the $F_{1}$ measure with ABRF-1 solving the classification
problem. Table \ref{t:classif_ABRF-1-1} is similar to Tables
\ref{t:regression_1_2}-\ref{t:regression_1_3} obtained for the regression
problem. It can be seen from Table \ref{t:classif_ABRF-1-1} that ABRF-1 mainly
shows outperforming results. However, there are several datasets (Eeg,
Ionosphere, Phishing) which demonstrate worse results for ABRF-1. This may be
due to use of the Euclidean distance between the one-hot vector $\mathbf{h}%
_{s}$ and the class probability distribution $\sum_{k=1}^{T}\alpha\left(
\mathbf{x}_{s},\mathbf{A}_{k}(\mathbf{x}_{s}\mathbf{)},\mathbf{w}\right)
\mathbf{p}(\mathbf{x}_{s})$ in the loss function (\ref{RF_Att_59}). It is
well-known that this comparison of the probability distributions is not the
best way in classification. However, different distances between probability
distributions, for example, the Kullback-Leibler divergence, complicates the
optimization problem (\ref{RF_Att_60}), which becomes non-quadratic.
Therefore, the Euclidean distance is used to save properties of the ABRF-1 simplicity.%

\begin{table}[tbp] \centering
\caption{The $F_1$ measure for comparison of models (the RF, the Softmax model, ABRF-1) trained on classification datasets under two conditions of training}%
\begin{tabular}
[c]{ccccccccc}\hline
& \multicolumn{4}{c}{Condition 1} & \multicolumn{4}{c}{Condition 2}\\\hline
Data set & $\epsilon$ & RF & Softmax & ABRF-1 & $\epsilon$ & RF & Softmax &
ABRF-1\\\hline
Diabet & $1$ & $0.626$ & $0.626$ & $\mathbf{0.628}$ & $0.75$ & $0.669$ &
$0.666$ & $\mathbf{0.672}$\\\hline
Eeg & $1$ & $0.618$ & $0.615$ & $\mathbf{0.672}$ & $0.75$ & $\mathbf{0.893}$ &
$0.889$ & $0.892$\\\hline
Haberman & $1$ & $0.425$ & $0.426$ & $\mathbf{0.578}$ & $0.5$ & $0.584$ &
$0.565$ & $\mathbf{0.594}$\\\hline
Ionosphere & $0.5$ & $\mathbf{0.912}$ & $0.910$ & $0.899$ & $0.5$ & $0.917$ &
$0.922$ & $\mathbf{0.926}$\\\hline
Seeds & $0.25$ & $0.897$ & $0.919$ & $\mathbf{0.929}$ & $0.25$ & $0.911$ &
$0.918$ & $\mathbf{0.923}$\\\hline
Seismic & $0.75$ & $\mathbf{0.483}$ & $\mathbf{0.483}$ & $\mathbf{0.483}$ &
$0.75$ & $0.483$ & $0.483$ & $\mathbf{0.486}$\\\hline
Soybean & $0.25$ & $0.989$ & $0.989$ & $\mathbf{1.000}$ & $0.5$ & $0.642$ &
$0.782$ & $\mathbf{0.978}$\\\hline
TAE & $0.75$ & $0.486$ & $0.471$ & $\mathbf{0.514}$ & $0.25$ & $\mathbf{0.485}%
$ & $\mathbf{0.485}$ & $0.483$\\\hline
TTTE & $0.5$ & $0.486$ & $0.570$ & $\mathbf{0.614}$ & $0.5$ & $0.792$ &
$0.824$ & $\mathbf{0.882}$\\\hline
Phishing & $0.25$ & $\mathbf{0.584}$ & $0.578$ & $0.577$ & $1$ & $0.607$ &
$0.606$ & $\mathbf{0.812}$\\\hline
Wholesale & $0.75$ & $\mathbf{0.278}$ & $\mathbf{0.278}$ & $\mathbf{0.278}$ &
$0$ & $\mathbf{0.278}$ & $\mathbf{0.278}$ & $\mathbf{0.278}$\\\hline
\end{tabular}
\label{t:classif_ABRF-1-1}%
\end{table}%

Table \ref{t:classif_ABRF-2-1} shows the $F_{1}$ measures obtained for ABRF-2
and ABRF-3 under Condition 2. We do not show results corresponding to
Condition 1 because they are similar to results presented for Condition 2.
Values of $\epsilon$ are shown in Table \ref{t:classif_ABRF-2-1} for the
ABRF-3 model. One can see from Table \ref{t:classif_ABRF-2-1} that the results
correlate with the same results for regression (see Tables
(\ref{t:regression_ABRF-2})-(\ref{t:regression_ABRF-2-3})), i.e., the ABRF-2
model does not show any sufficient improvement in comparison with the original
RF. In contrast to results presented in Table \ref{t:classif_ABRF-2-1} for
ABRF-2 and even in Table \ref{t:classif_ABRF-1-1} for ABRF-1, the ABRF-3 model
provides outperforming results as it is shown in Table
\ref{t:classif_ABRF-2-1}. However, it is interesting to point out that ABRF-3
is actually reduced to ABRF-1 for several datasets when $\epsilon=1$. In this
case, the term with the softmax function is not used, and the attention
weights are entirely determined by the $\epsilon$-contamination model. United
results obtained for the case of the ERT usage under Condition 2 are presented
in Table \ref{t:classif_ABRF-2-3}. It can be again seen from Table
\ref{t:classif_ABRF-2-3} that the proposed attention-based models outperform
the ERT and the ERT with the softmax function without its training for all
considered datasets.

In sum, results of numerical experiments for classification generally coincide
with the results obtained for regression. Analyzing these results, we can
conclude that the ABRF-3 is comparable with ABRF-1 for regression as well as
for classification. Moreover, numerical experiments show that ABRF-3 provides
results which are slightly better than results of ABRF-1. However, ABRF-1 is
computationally much more simpler than the ABRF-3 model because it is based on
solving the linear or quadratic optimization problems whereas ABRF-3 requires
to solve the complex optimization problems for training.%

\begin{table}[tbp] \centering
\caption{The $F_1$ measure for comparison of models (the RF, the Softmax model, ABRF-2, ABRF-3) trained on classification datasets under Condition 2 of training}%
\begin{tabular}
[c]{cccccc}\hline
Data set & RF & Softmax & ABRF-2 & $\epsilon$ & ABRF-3\\\hline
Diabet & $0.669$ & $0.666$ & $0.610$ & $1$ & $\mathbf{0.671}$\\\hline
Eeg & $0.893$ & $0.889$ & $\mathbf{0.909}$ & $0$ & $\mathbf{0.909}$\\\hline
Haberman & $0.584$ & $0.566$ & $0.536$ & $0.5$ & $\mathbf{0.599}$\\\hline
Ionosphere & $0.917$ & $0.922$ & $0.889$ & $1$ & $\mathbf{0.924}$\\\hline
Seeds & $0.911$ & $\mathbf{0.918}$ & $0.892$ & $1$ & $0.911$\\\hline
Seismic & $0.483$ & $0.483$ & $\mathbf{0.502}$ & $0$ & $\mathbf{0.502}%
$\\\hline
Soybean & $0.642$ & $0.782$ & $0.926$ & $1$ & $\mathbf{0.978}$\\\hline
TAE & $0.485$ & $0.485$ & $0.487$ & $0.5$ & $\mathbf{0.520}$\\\hline
TTTE & $0.792$ & $0.824$ & $0.891$ & $0.25$ & $\mathbf{0.900}$\\\hline
Phishing & $0.607$ & $0.606$ & $0.790$ & $1$ & $\mathbf{0.812}$\\\hline
Wholesale & $0.278$ & $0.278$ & $0.278$ & $0.25$ & $\mathbf{0.284}$\\\hline
\end{tabular}
\label{t:classif_ABRF-2-1}%
\end{table}%
%

\begin{table}[tbp] \centering
\caption{The $F_1$ measure for comparison of models (the ERT, the Softmax model, ABRF-1, ABRF-2, ABRF-3) trained on classification datasets under Condition 2 of training}%
\begin{tabular}
[c]{cccccc}\hline
Data set & ERT & Softmax & ABRF-1 & ABRF-2 & ABRF-3\\\hline
Diabet & $0.664$ & $0.664$ & $\mathbf{0.669}$ & $0.646$ & $0.664$\\\hline
Eeg & $0.355$ & $0.355$ & $0.722$ & $0.733$ & $\mathbf{0.736}$\\\hline
Haberman & $0.426$ & $0.426$ & $0.550$ & $\mathbf{0.561}$ & $\mathbf{0.561}%
$\\\hline
Ionosphere & $0.911$ & $0.911$ & $\mathbf{0.914}$ & $0.881$ & $0.912$\\\hline
Seeds & $0.904$ & $0.921$ & $\mathbf{0.938}$ & $0.933$ & $0.933$\\\hline
Seismic & $0.483$ & $0.483$ & $0.483$ & $\mathbf{0.507}$ & $\mathbf{0.507}%
$\\\hline
Soybean & $0.847$ & $0.896$ & $\mathbf{0.964}$ & $0.949$ & $0.949$\\\hline
TAE & $0.511$ & $0.502$ & $\mathbf{0.517}$ & $0.498$ & $\mathbf{0.517}%
$\\\hline
TTTE & $0.890$ & $0.924$ & $\mathbf{0.937}$ & $0.903$ & $0.931$\\\hline
Phishing & $0.609$ & $0.609$ & $0.816$ & $\mathbf{0.849}$ & $\mathbf{0.849}%
$\\\hline
Wholesale & $\mathbf{0.278}$ & $\mathbf{0.278}$ & $\mathbf{0.278}$ & $0.277$ &
$\mathbf{0.278}$\\\hline
\end{tabular}
\label{t:classif_ABRF-2-3}%
\end{table}%

\section{Concluding and discussion remark}

New models of the attention-based RF have been proposed. They inherit the best
properties of the attention mechanism and allow us to avoid using neural
networks. Moreover, some models allow us to avoid using the gradient-based
algorithms for computing the trainable parameters of the attention because the
quadratic or linear programming can be applied to training. The idea to avoid
differentiable nonlinear modules in machine learning models has been
highlighted by Zhou and Feng \cite{Zhou-Feng-2019} where it has been realized
in the so-called deep forests. We also tried to realize this idea in the
attention mechanism by applying it to RFs and by introducing the Huber's
$\epsilon$-contamination model in ABRF-1. At the same time, we did not deny
applying the gradient-based algorithm as an efficient tool for solving
optimization problems, and ABRF-2, ABRF-3 have also demonstrated desirable
results. However, they tend to overfitting for some datasets and parameters.
Therefore, we think that the most interesting model is ABRF-1.

Let us point out \textit{advantages} of the proposed models. First, the ABRF-1
models are simply trained by solving the standard quadratic or linear
optimization problems. Second, the proposed usage of the Huber's $\epsilon
$-contamination model extends the set of weight functions in attention models.
Third, the proposed models are flexible and can be simply modified. For
example, different procedures for computing values $\mathbf{A}_{k}%
(\mathbf{x)}$ can be proposed and studied, different kernels can be used
instead of the softmax. Fourth, the attention-based approach allows us to
significantly improve predictions, and numerical experiments have illustrated
this improvement on several datasets. Fifth, the results are interpretable
because the attention weights show which decision trees have the largest
contribution into predictions. Sixth, when new training instances appear after
building the RF, we do not need to rebuild trees because it is enough to train
weights of trees. Finally, the models are implementation of the attention
mechanism on RFs without neural networks.

The following \textit{disadvantages} should be also pointed out. First, the RF
is not the best classifier or regressor. It perfectly deals with tabular data,
but many other types of data, for example, images, may result unsatisfactory
predictions in comparison with, for example, neural networks. Second, the
ABRF-1 model is very simple from the computation point of view. However, it
has the tuning parameter $\epsilon$ which requires to solve many optimization
problems in order to get the best results. Moreover, we have to take into
account the temperature parameter $\tau$ of the softmax function. Third, most
attention models directly provide weights of instances. The proposed ABRF
models compute weights of trees. It should be noted that the weights of
instances can be calculated through the weights of trees, but an additional
analysis has to be performed for doing that.

Numerical experiments have demonstrated that the attention models can
significantly improve original RFs. Moreover, many extensions and new models
can be developed due to flexibility of the ABRF models. First of all, it is
interesting to extend the proposed approach to the gradient boosting machine
\cite{Friedman-2002} which is based on decision trees as week learners. It is
also interesting to extend the approach on deep forests \cite{Zhou-Feng-2019}.
These are directions for further research. Another interesting direction for
research is to investigate various functions instead of the softmax which is
used in the proposed models. It is obvious that kernel functions should be
used. Choice of the kernel functions, which lead to simple computations and
outperforming prediction results, is a direction for further research. In
models ABRF-1 and ABRF-3, objective functions have been optimized over the
whole set of weights, i.e., over the unit simplex. However, the set of weights
can be restricted under various conditions. This restriction may prevent from
overfitting. Moreover, there are other statistical contamination models
\cite{Walley91} which could be incorporated into the ABRF models. These ideas
can be also viewed as directions for further research. Finally, we have
considered the distance between vectors $\mathbf{x}_{s}$ and $\mathbf{A}%
_{k}(\mathbf{x}_{s}\mathbf{)}$ inside one leaf where the instance
$\mathbf{x}_{s}$ falls into. However, it may be useful to extend this distance
definition and to study some weighted sum of distances between $\mathbf{x}%
_{s}$ and $t$ vectors $\mathbf{A}_{k}(\mathbf{x}_{s}\mathbf{)}$ which are
defined for $t$ nearest neighbor leaves. Moreover, there are different
definitions of vector $\mathbf{A}_{k}(\mathbf{x}_{s}\mathbf{)}$ itself, for
example, we can use the median instead of the mean value. This is also an
interesting direction for further research.

\section*{Acknowledgement}

This work is supported by the Russian Science Foundation under grant 21-11-00116.

\bibliographystyle{unsrt}
\bibliography{Attention,Boosting,Deep_Forest,Expl_Attention,Explain,Explain_med,Imprbib,Lasso,MIL,MYBIB,MYUSE}

\end{document}